\def\paperTitle{Binary Hypothesis Testing for Softmax Models and Leverage Score Models}

\def\paperAuthor{
Yuzhou Gu
\and
Zhao Song\thanks{\texttt{magic.linuxkde@gmail.com}. University of California, Berkeley.}
\and
Junze Yin
}

\ifdefined\isarxiv
\documentclass[11pt]{article}

\usepackage[numbers]{natbib}

\else

\documentclass{article}

\usepackage{microtype}
\usepackage{graphicx}
\usepackage{subfigure}
\usepackage{booktabs} 

\usepackage{hyperref}

\usepackage[accepted]{icml2025}

\usepackage{amsmath}
\usepackage{amssymb}
\usepackage{mathtools}
\usepackage{amsthm}

\theoremstyle{plain}
\newtheorem{theorem}{Theorem}[section]
\newtheorem{proposition}[theorem]{Proposition}
\newtheorem{lemma}[theorem]{Lemma}
\newtheorem{corollary}[theorem]{Corollary}
\theoremstyle{definition}
\newtheorem{definition}[theorem]{Definition}
\newtheorem{assumption}[theorem]{Assumption}
\theoremstyle{remark}
\newtheorem{remark}[theorem]{Remark}

\usepackage[textsize=tiny]{todonotes}

\icmltitlerunning{Binary Hypothesis Testing for Softmax Models and Leverage Score Models}

\fi

\usepackage{amsmath}
\usepackage{amsthm}
\usepackage{amssymb}
\usepackage{algorithm} 
\usepackage{algpseudocode}
\usepackage{grffile}
\usepackage{wrapfig,epsfig}
\usepackage{url}
\usepackage{xcolor}
\usepackage{epstopdf}
\usepackage{bbm}
\usepackage{dsfont}

\allowdisplaybreaks

\ifdefined\isarxiv

\usepackage{tikz}
\usepackage{hyperref}  
\hypersetup{colorlinks=true,citecolor=blue,linkcolor=blue} 
\usetikzlibrary{arrows}
\usepackage[margin=1in]{geometry}

\else

\fi

\theoremstyle{plain}
\ifdefined\isarxiv
\newtheorem{theorem}{Theorem}[section]
\newtheorem{lemma}[theorem]{Lemma}
\newtheorem{definition}[theorem]{Definition}

\newtheorem{corollary}[theorem]{Corollary}

\else

\fi

\newcommand{\N}{\mathcal{N}}
\newcommand{\R}{\mathbb{R}}

\renewcommand{\d}{\mathrm{d}}

\DeclareMathOperator*{\E}{{\mathbb{E}}}

\DeclareMathOperator{\diag}{diag}
\DeclareMathOperator{\Diag}{Diag}

\DeclareMathOperator{\tr}{tr}

\DeclareMathOperator{\TV}{TV}
\DeclareMathOperator{\Var}{Var}
\DeclareMathOperator{\op}{op}
\DeclareMathOperator{\Softmax}{\mathtt{SoftMax}}
\DeclareMathOperator{\Leverage}{\mathtt{Leverage}}

\makeatletter
\newcommand*{\RN}[1]{\expandafter\@slowromancap\romannumeral #1@}
\makeatother

\usepackage{lineno}

\begin{document}

\ifdefined\isarxiv

\date{}

\title{\paperTitle}
\author{\paperAuthor}

\else

\twocolumn[
\icmltitle{Binary Hypothesis Testing for Softmax Models and Leverage Score Models}



\icmlsetsymbol{equal}{*}

\begin{icmlauthorlist}
\icmlauthor{Yuzhou Gu}{ias}
\icmlauthor{Zhao Song}{ucb}
\icmlauthor{Junze Yin}{bu}
\end{icmlauthorlist}

\icmlaffiliation{bu}{Boston University}
\icmlaffiliation{ias}{Institute for Advanced Study}
\icmlaffiliation{ucb}{University of California Berkeley}
 
\icmlcorrespondingauthor{Zhao Song}{magic.linuxkde@gmail.com}

\icmlkeywords{Machine Learning, ICML}

\vskip 0.3in
]



\printAffiliationsAndNotice{}  

\fi

\ifdefined\isarxiv
\begin{titlepage}
  \maketitle
  \begin{abstract}
Softmax distributions are widely used in machine learning, including Large Language Models (LLMs), where the attention unit uses softmax distributions. We abstract the attention unit as the softmax model, where given a vector input, the model produces an output drawn from the softmax distribution (which depends on the vector input). We consider the fundamental problem of binary hypothesis testing in the setting of softmax models. That is, given an unknown softmax model, which is known to be one of the two given softmax models, how many queries are needed to determine which one is the truth? We show that the sample complexity is asymptotically $O(\epsilon^{-2})$ where $\epsilon$ is a certain distance between the parameters of the models. 
Furthermore, we draw an analogy between the softmax model and the leverage score model, an important tool for algorithm design in linear algebra and graph theory. The leverage score model, on a high level, is a model which, given a vector input, produces an output drawn from a distribution dependent on the input. We obtain similar results for the binary hypothesis testing problem for leverage score models.

  \end{abstract}
  \thispagestyle{empty}
\end{titlepage}

{\hypersetup{linkcolor=black}
\tableofcontents
}
\newpage

\else
\begin{abstract}
Softmax distributions are widely used in machine learning, including Large Language Models (LLMs), where the attention unit uses softmax distributions. We abstract the attention unit as the softmax model, where given a vector input, the model produces an output drawn from the softmax distribution (which depends on the vector input). We consider the fundamental problem of binary hypothesis testing in the setting of softmax models. That is, given an unknown softmax model, which is known to be one of the two given softmax models, how many queries are needed to determine which one is the truth? We show that the sample complexity is asymptotically $O(\epsilon^{-2})$ where $\epsilon$ is a certain distance between the parameters of the models. 
Furthermore, we draw an analogy between the softmax model and the leverage score model, an important tool for algorithm design in linear algebra and graph theory. The leverage score model, on a high level, is a model which, given a vector input, produces an output drawn from a distribution dependent on the input. We obtain similar results for the binary hypothesis testing problem for leverage score models.
\end{abstract}

\fi

\section{Introduction}
\label{sec:intro}

In transforming various aspects of people's lives, large language models (LLMs) have exhibited tremendous potential. In recent years, numerous content learning and LLMs have been developed, including notable models such as Adobe Firefly, Microsoft 365 Copilot \citep{s23}, Adobe Photoshop, and Google's Meena chatbot \citep{r20}, along with the GPT series, the DeepSeek series, Google's Gemini series and others \citep{rns+18,rwc+19, dclt18,rwc+19,ydy+19,bmr+20,cha22,o23,gemini,deepseekv3}. These models, together with those built upon them, have demonstrated significant prowess across diverse fields.
The robustness and vitality of their development are attested to by the widespread integration of LLMs. In the realm of Natural Language Processing (NLP), evaluations by \cite{lbl+22, lbr+23, cpk+23, bcl+23} center around natural language understanding, while \cite{wlj+23, qzz+23, pd23, chbp23, cwj+23} delve into natural language generation. LLMs have found applications in diverse fields, including both social science and science \citep{ggl+23,dgg23, f23, nkl+23}, medical applications \citep{clbj23, jgp+23}, cybersecurity~\cite{sdj+23,xwl+24,lsj25,hst+25,daw25} and engineering \citep{pmm+23, sm+23, bce+23, lxwz23}, showcasing their potent capabilities.
A consistent theme among these models is the adoption of the transformer architecture, a proven and highly efficient framework. The prevailing prevalence of models like ChatGPT \citep{o23} further underscores the transformative impact of this architecture.

However, there is a crucial problem with LLMs: their training costs and uncertainty regarding their inference ability in different parts of the whole. Understanding how different domains work is important in retrieval argument generation (RAG) \citep{sww+23, zb24, sz24}, as well as sparsity for LLMs by identifying the ability domain in the model which is important in solving the problem above. Then a question arose:
\begin{center}
    {\it Can we distinguish different ability parts of large language models by limited parameters sampling?}
\end{center}
We take an initial step toward addressing this question from a theoretical perspective. As we delve deeper into LLMs, the softmax mechanism is found to play an important role in the computation of self-attention. Thus, it is imperative to study how the self-attention mechanism works, why it contributes significantly to the impressive capabilities of LLMs, and what role it plays are still not fully understood.

Therefore, in this work, we want to explore the mechanism of softmax distribution from a binary hypothesis testing perspective. By delving into the intricacies of the softmax formulation, we explore which parameters are important by explaining how the softmax can be distinguished from each other. By delving into this idea, we can determine how many parameters are important in the inference of transformers \citep{vsp+17}.
In continuation of the paper and drawing upon a formulation similar to softmax, we also direct our attention to the distribution of leverage scores. Much like softmax, the leverage score is a distribution parameterized by a matrix. Both softmax and leverage score can be treated as functions of distribution within this context. Importantly, resembling softmax, leverage score assumes significance across various fields. 
Leverage scores have demonstrated their significant utility in both linear algebra and graph theory. In the field of graph theory, researchers have extensively explored the application of leverage scores in various areas such as the generation of random spanning trees \citep{s18}, max-flow problems \citep{ds08, m13, m16, ls20}, maximum matching \citep{bln+20, lsz20}, and graph sparsification \citep{s11}. Many studies have delved into the deep exploration of leverage scores, showcasing their effectiveness in optimization tasks such as linear programming \citep{ls14,cls19,s19,lsz19,blss20,sy21,lsz+22,syz23,qszz23,gsz25}, cutting-plane methods \citep{v89, lsw15, jlsw20}, semi-definite programming \citep{jkl+20,hjs+22,ckp+23,gsyz23}, and the approximation of the John Ellipsoid \citep{ccly19,syyz22,lls+24,lsy24}. These applications underscore the importance of leverage scores in the context of theory of computer science and linear algebra.
Based on the analysis provided, both the leverage score and softmax computation are parameterized by a single matrix. Given the significance of the application of softmax and computation, understanding the influence on parameter behavior becomes crucial. Hence, we delve into this inquiry by differentiating the model through parameter sampling and discussing how the number of samples affects the distinguishing ability.

A softmax model is parameterized by a matrix $A \in \R^{n\times d}$, and denoted $\Softmax_A$. Given $x\in \R^d$, the model outputs an element $i\in [n]$ with probability
$
    p_i = \langle \exp(Ax), {\bf 1}_n\rangle^{-1} \exp(Ax)_i.
$
In the binary hypothesis testing problem, we are given access to a softmax model which is either $\Softmax_A$ or $\Softmax_B$. We have query access to the model, that is, we can feed the model an input $x\in \R^d$, and it will produce an output. The goal is to determine whether the model is $\Softmax_A$ or $\Softmax_B$, using the fewest number of queries possible.
We can similarly define the question for leverage score models. A leverage score model is parameterized by a matrix $A\in \R^{n\times d}$, and denoted $\Leverage_A$. Given input $s\in (\R \backslash \{0\})^n$, the model returns an element $i\in [n]$ with probability 
$
p_i = (A_s (A_s^\top A_s)^{-1} A_s^\top)_{i,i}/d,
$
where $A_s = S^{-1}A$, and $S=\Diag(s)$ is the diagonal matrix with diagonal $s$. We define the binary hypothesis testing problem for leverage score models similarly to the softmax case. 

\subsection{Main Result.}
We state informal versions of our main results.
\begin{theorem}[Informal statement of Theorem~\ref{thm:softmax-testing-lower-bound} and Theorem~\ref{thm:softmax-testing-upper-bound-2}]\label{thm:softmax}
Consider the binary hypothesis testing problem with two softmax models $\Softmax_A$ and $\Softmax_B$. We have 1). if $\|B-A\|_{2\to \infty} \le \epsilon$, then any successful algorithm uses $\Omega(\epsilon^{-2})$ queries (Lower bound), and 2). if $B = A + \epsilon M$ for some small $\epsilon$ then the hypothesis testing problem can be solved in $O(\epsilon^{-2} \nu)$ queries, where $\nu$ depends on $A$ and $M$ (Upper bound).
\end{theorem}

\begin{theorem}[Informal statement of Theorem~\ref{thm:leverage-testing-lower-bound} and Theorem~\ref{thm:leverage-testing-upper-bound-2}]\label{thm:leverage}
Consider the binary hypothesis testing problem with  two leverage score models $\Leverage_A$ and $\Leverage_B$. We have 1). if $\sum_{i\in [n]} \|B_{i,*}^\top B_{i,*} - A_{i,*}^\top A_{i,*}\|_{\op} \le \epsilon$, then any successful algorithm uses $\Omega(\epsilon^{-1})$ queries (Lower bound), and 2). if $B = A + \epsilon M$ for some small $\epsilon$ then the hypothesis testing problem can be solved in $O(\epsilon^{-2} \nu)$ queries, where $\nu$ depends on $A$ and $M$ (Upper bound).
\end{theorem}

\subsection{Related Work}

\paragraph{Theoretical LLMs}

Several investigations \citep{clmy21, llh+23, ryw+19, hm19} have concentrated on theoretical analyses concerning LLMs.
The algorithm presented by \cite{clmy21}, named ZO-BCD, introduces a novel approach characterized by advantageous overall query complexity and reduced computational complexity in each iteration. The work by \cite{llh+23} introduces Sophia, a straightforward yet scalable second-order optimizer. Sophia demonstrates adaptability to curvature variations across different parameter regions, a feature particularly advantageous for language modeling tasks with strong heterogeneity. Importantly, the runtime bounds of Sophia are independent of the condition number of the loss function. 

Studies by \cite{wwz+22,ll21,ddh+21,byks22,hbkg23,xqp+22} investigate the knowledge and skills of LLMs.
In the realm of optimization for LLMs, \cite{kmh+20,rsm+23,llh+23,cls24} have delved into this domain. 
Demonstrating the effectiveness of pre-trained models in localizing knowledge within their feed-forward layers, both \cite{hbkg23} and \cite{mbab22} contribute valuable insights to the field. 
The exploration of distinct "skill" neurons and their significance in soft prompt-tuning for language models is a central theme in the analysis conducted by \cite{wwz+22}, building upon the groundwork laid out in a prior discussion by \cite{ll21}. The activation of skill neurons and their correlation with the expression of relevant facts is a focal point in the research presented by \cite{ddh+21}, particularly in the context of BERT. In contrast, the work of \cite{byks22} takes an entirely unsupervised approach, leveraging the internal activations of a language model to extract latent knowledge. Next, the investigation by \cite{lyb+22} sheds light on the sparsity observed in feedforward activations of large trained transformers, uncovering noteworthy patterns in their behavior.
In addition to the above, \cite{clmy21,mgn+23,dlms23,zhl+23} explore zeroth order algorithms for LLMs, and \cite{hsw+22,hsk+24,c24} explore parameter-efficient fine-tuning of LLMs.

Several recent works have revealed the inherent limitations of LLMs from a theoretical perspective. These limitations are grounded in empirical benchmarks indicating that LLMs may face difficulties in simple tasks~\cite{zzc+23,ghh+25,cgh+25,ghs+25_physical}. An important line of research~\cite{as23,as24_iclr,as24_neurips,as25_rank,as25_rope} has theoretically shown that the forward and backward passes of Transformers cannot be approximated with low error in truly subquadratic time, unless the attention weights are sufficiently small in each entry. 
Beyond the significant success of circuit complexity in showing that many neural architectures~\cite{kll+25_tc,cll+25_mamba,lls+25} may belong to a weak class of logical circuits, such as $\mathsf{TC}^0$, and may fail to solve harder problems, these results have effectively explained LLMs' limitations~\cite{wcm+22,llzm24,cll+24_rope}. For example, they show that Transformers cannot solve arithmetic formula evaluation and that CoT may enhance LLMs' circuit complexity bounds. 
Recent works have also gone beyond numerical approximation and model expressiveness, finding that under gradient descent training, certain types of simple Boolean function problems may be difficult for LLMs to learn, such as the parity problem~\cite{ks25}, the majority problem~\cite{cssz25}, and simpler and/or problems~\cite{hzs+25}.

\paragraph{Leverage Scores} 
Given $A \in \mathbb{R}^{n \times d}$ and  $i \in [n]$, $a_i$ represents the $i$-th row of matrix $A$. We use  $\sigma_i(A) = a_i^\top (A^\top A)^\dagger a_i$ to denote the leverage score for the $i$-th row of matrix $A$. The concept of leverage score finds extensive applications in the domains of machine learning and linear algebra. In numerical linear algebra and graph theory, leverage scores serve as fundamental tools.
In the context of matrices, both the tensor CURT decomposition \citep{swz19} and the matrix CUR decomposition \citep{bw14, swz17, swz19} heavily rely on leverage scores. In optimization, areas such as linear programming \citep{ls14, blss20}, the approximation of the John Ellipsoid \citep{ccly19}, cutting-plane methods \citep{v89, lsw15, jlsw20}, and semi-definite programming \citep{jkl+20} incorporate leverage scores.
Within graph theory applications, leverage scores play a crucial role in max-flow problems \citep{ds08, m13, m16, ls20}, maximum matching \citep{bln+20, lsz20}, graph sparsification \citep{s11}, and the generation of random spanning trees \citep{s18}. 
Several studies, such as \cite{ss11, DMIMW12, cw13}, focus on the approximation of leverage scores. Simultaneously, Lewis weights, serving as a generalization of leverage scores, are explored in depth by \cite{blm89, cp15}. 
Notably, leverage scores may also motivate efficient computational methods in machine learning, including regression problems~\cite{psw17,swy+19,gst22,syyz23_linf,s19}, deep neural networks~\cite{lls+20,zha+21}, graph neural networks~\cite{ssz23_gnn,zha24}, recommender systems~\cite{hldt23,zxf+24}, active learning~\cite{scmw24,gtx+24}, discrepancy minimization~\cite{dlsw25} and Fourier transform~\cite{lxz25}.

\paragraph{Hypothesis Testing}
Hypothesis testing is a central problem in statistics, and has many applications in machine learning in recent years~\citep{lxl+20,glz+21,lxls21,xzl+22,syz+25}. In hypothesis testing, two (or more) hypotheses about the truth are given and an algorithm needs to distinguish which hypothesis is true. 
The most classic testing problem is the binary hypothesis testing.
In this problem, two distributions $P_0$ and $P_1$ are given, and there is an unknown distribution $P$ which is either $P_0$ or $P_1$. The goal is to distinguish whether $P=P_0$ or $P=P_1$ by drawing samples from $P$. This problem is well-studied, with \cite{neyman1933ix} giving tight characterization of the possible error regions in terms of the likelihood ratio. It is known that the asymptotic sample complexity of binary hypothesis testing for distributions is given by $\Theta(H^{-2}(P_0,P_1))$, where $H$ denotes the Hellinger distance, see e.g., \cite{polyanskiy2023information}.
There are other important kinds of hypothesis testing problems. In the goodness-of-fit testing problem, a distribution $Q$ is given, and there is an unknown distribution $P$ which is known to be either equal to $Q$ or far away from $Q$. The goal is to distinguish which is the true by drawing samples from $P$. In the two-sample testing problem, two unknown distributions $P$ and $Q$ are given, and it is known that either $P=Q$ or $P$ and $Q$ are far away from each other. The goal is to distinguish which is true by drawing samples from $P$ and $Q$.
For these problems there are no simple general characterization as in the binary hypothesis testing. However, for reasonable classes of distributions such as Gaussian distributions or distributions on discrete spaces, a lot of nice results are known \citep{ingster1987minimax,ingster1982minimax,goldreich2011testing,valiant2017automatic,chan2014optimal,arias2018remember,li2019optimality}.
We are not aware of any previous work that studies hypothesis testing problems for the class of softmax models or leverage score models.

\paragraph{Roadmap.}
In Section~\ref{sec:preli}, we introduce notation and concepts related to information theory and hypothesis testing. Our results are presented in Section~\ref{sec:softmax} and Section~\ref{sec:leverage}: Section~\ref{sec:softmax} establishes upper and lower bounds on the sample complexity for distinguishing two different softmax models, and Section~\ref{sec:leverage} delves into the case of leverage scores.
We present several additional remarks for our main results in Section~\ref{sec:discussion}.
We conclude and make further discussions in Section~\ref{sec:conclusion}.

\section{Preliminaries}\label{sec:preli}
In Section~\ref{sec:notation}, we define several basic notations. In Section~\ref{sec:preli:information_theory}, we provide definitions related to information theory. In Section~\ref{sec:preli:hypothesis_testing}, we provide backgrounds about hypothesis testing. In Section~\ref{sec:preli:softmax}, we provide definition of softmax model. In Section~\ref{sec:preli:leverage}, we provide definition of leverage score model.

\subsection{Notation}\label{sec:notation}
Given $x \in \R^n$, we use $\| x \|_p$ to denote $\ell_p$ norm of $x$, where $\| x \|_0 = \sum_{i=1}^n \mathbbm{1}(x_i \neq 0)$, $\|x\|_1 := \sum_{i=1}^n |x_i|$ ($\ell_1$ norm), $\|x\|_2 := (\sum_{i=1}^n x_i^2)^{1/2}$ ($\ell_2$ norm), and $\|x\|_\infty := \max_{i\in [n]} |x_i|$ ($\ell_\infty$ norm). For a square matrix, $\tr[A]$ is used to represent the trace of $A$.
Given $1\le p \le \infty$ and $1 \le q\le \infty$, $\|A\|_{p\to q}$ represents the $p$-to-$q$ operator norm
$\|A\|_{p\to q} = \sup_{x: \|x\|_p\le 1} \|A x\|_q$.
In particular, $\|A\|_{2\to \infty} = \max_{i\in [n]} \|A_{i,*}\|_2$.
For $x\in \R^n$, let $\Diag(x)\in \R^{n\times n}$ denote the diagonal matrix with diagonal $x$.
For square matrix $A\in \R^{n\times n}$, let $\diag(A)\in \R^n$ denote the diagonal of $A$.
For a non-negative integer $n$, let $[n]$ denote the set $\{1,\ldots,n\}$.
For a sequence $X_1,\ldots,X_m$ of random variables, we use $X^m$ to denote the whole sequence $(X_1,\ldots,X_m)$.

\subsection{Information Theory}\label{sec:preli:information_theory}

\begin{definition}[TV distance]\label{def:tv_distance}
For two distributions $P, Q$ on the same measurable space, their total variation (TV) distance is
\begin{align*}
    \TV(P, Q) = \frac 12 \int \left|P(\d x) - Q(\d x)\right |.
\end{align*}
In particular, if $P$ and $Q$ are on the discrete space $[n]$ and $P=(p_1,\ldots,p_n)$, $Q=(q_1,\ldots,q_n)$, then
\begin{align*}
    \TV(P, Q)) = \frac 12 \sum_{i=1}^n |p_i-q_i |.
\end{align*}
\end{definition}

\begin{definition}[Hellinger distance]\label{def:hellinger_distance}
For two distributions $P,Q$ on the same measurable space, their squared Hellinger distance is
\begin{align*}
H^2(P,Q) = \frac{1}{2} \int (\sqrt{P(\d x)}-\sqrt{Q(\d x)})^2.
\end{align*}
In particular, if $P$ and $Q$ are on the discrete space $[n]$ and $P=(p_1,\ldots,p_n)$, $Q=(q_1,\ldots,q_n)$, then 
\begin{align*}
    H^2(P,Q) 
    = & ~ \frac{1}{2} \sum_{i=1}^n (\sqrt{p_i} - \sqrt{q_i})^2 \\
    = & ~ 1 - \sum_{i=1}^n \sqrt{p_i q_i}.
\end{align*}
The Hellinger distance $H(P,Q)$ is the square root of the squared Hellinger distance $H^2(P,Q)$.
\end{definition}

We recall the following relationship between the Hellinger distance and the TV distance. For any distributions $P, Q$ on the same space, we have
\begin{align*}
    H^2(P, Q)\le \TV(P, Q) \le \sqrt 2 H(P, Q).
\end{align*}

\begin{definition}[Expectation and variance]\label{def:var}
Let $P$ be a distribution on a measurable space $\cal X$ and $f$ be a continuous function on $\cal X$. Then $\E_P[f]$ is the expectation of $f$ under $P$ and $\Var_P(f)$ is the variance of $f$ under $P$.
In particular, if ${\cal X}=[n]$, $P=(p_1,\ldots,p_n)\in \R^n$, and $x\in \R^n$, then
\begin{align*}
\E_P[x] = \sum_{i=1}^n p_i x_i ~~ \mathrm{and} ~~ \Var_P(x) = \sum_{i=1}^n p_i ( x - \E_P[x])^2.
\end{align*} 
\end{definition}

\subsection{Hypothesis Testing}\label{sec:preli:hypothesis_testing}
We review the classic hypothesis testing problem for distributions.
\begin{definition}[Binary hypothesis testing for distributions]
Let $P_0, P_1$ be two distributions on the same space.
We have sample access to a distribution $P$, which is known to be either $P_0$ or $P_1$.
The goal is to determine whether $P=P_0$ or $P=P_1$, using as few samples as possible.
We say an algorithm successfully distinguishes $P_0$ and $P_1$ is at least $2/3$ under both hypotheses.
\end{definition}
In the above definition, the constant $2/3$ can be replaced by any constant $>1/2$, and the asymptotic sample complexity of the binary hypothesis testing problem does not change.
The reason is that if we have an algorithm that achieves success probability $\delta>\frac 12$, then we can run it independently a constant number of times and take the majority of the outputs. Thus, we can boost the success probability to an arbitrarily high constant.
A classic result in information theory states that the sample complexity of the binary hypothesis testing problem is determined by the Hellinger distance.
\begin{lemma}[e.g., \cite{polyanskiy2023information}] \label{lem:binary-hypo-test-classic}
The sample complexity of the binary hypothesis testing problem for distributions is $\Theta(H^{-2}(P_0, P_1))$. That is, there is an algorithm that solves the problem using $O(H^{-2}(P_0,P_1))$ queries, and any algorithm that solves the problem uses $\Omega(H^{-2}(P_0,P_1))$ queries.
\end{lemma}

\subsection{Softmax Model}\label{sec:preli:softmax}
\begin{definition}[Softmax model]\label{def:softmax_model}
The \emph{softmax model} $\Softmax_A$ associated with $A\in \R^{n\times d}$ is a model such that on input $x\in \R^d$, it outputs a sample $y\in [n]$ from the distribution $\Softmax_A(x)$, defined as follows: the probability mass of $i\in [n]$ is equal to $\langle \exp(Ax), {\bf 1}_n \rangle^{-1} \exp( Ax )_i$.
\end{definition}
Note that $\sum_{i=1}^n\langle \exp(Ax), {\bf 1}_n \rangle^{-1} \exp( Ax )_i = 1$, so the above definition gives a valid distribution.

\begin{definition}[Binary hypothesis testing for softmax models]\label{def:binary_hypothesis_testing}
Let $A,B\in \R^{n\times d}$ be two matrices.
Let $P_0=\Softmax_{A}, P_1=\Softmax_{B}$ be two softmax models.
Let $P$ be the softmax model which is either $P_0$ or $P_1$.
In each query, we can feed $x\in \R^d$ into $P$, and retrieve a sample $y\in [n]$ from $P(x)$.
The goal is to determine if the model $P$ is $P_0$ or $P_1$ in as few samples as possible.
We say an algorithm successfully distinguishes $P_0$ and $P_1$, if the correctness probability is at least $2/3$ under both hypotheses.
\end{definition}

The above definition is valid. However, if we make no restrictions on the input $x$, then there would be undesirable consequences. For example, suppose $n=2$, $d=1$, $A = \begin{bmatrix}\epsilon \\ 0 \end{bmatrix}$, $B = \begin{bmatrix}0 \\ \epsilon \end{bmatrix}$ for some very small $\epsilon>0$. Because $A$ and $B$ are close to each other, we should expect it to be difficult to distinguish $\Softmax_A$ and $\Softmax_B$. However, if we allow any $x\in \R^d$ as input, then we could take $x$ to be a very large real number. Then $\Softmax_A(x)$ has almost all mass on $1\in [n]$, while $\Softmax_B(x)$ has almost all mass on $2\in [n]$, and we can distinguish the two models using only one query. To avoid this peculiarity, we assume that there is an energy constraint on $x$.

\begin{definition}[Energy constraint for softmax model] \label{def:softmax-constraint}
We assume that there is an \emph{energy constraint}, that is, input $x\in \R^n$ should satisfy $\|x\|_2 \le E$, for some given constant $E$.
\end{definition}
The energy constraint is a reasonable assumption in the context of LLMs and more generally neural networks, because of the widely used batch normalization technique \citep{ioffe2015batch}.

\subsection{Leverage Score Model}\label{sec:preli:leverage}

\begin{definition}[Leverage score model]
The \emph{leverage score model} $\Leverage_A$ associated with $A\in \R^{n\times d}$ is a model such that on input $s\in (\R\backslash \{0\})^{n}$, it outputs a sample $y \in [n]$ from the distribution $\Leverage_A(s)$, defined as follows: the probability mass of $i \in [n]$ is equal to 
\begin{align*}
\| (A_s^\top A_s)^{-1/2} (A_s)_{*,i} \|_2^2 / d =  ( A_s (A_s^\top A_s)^{-1} A_s^\top )_{i,i} / d,
\end{align*}
where $A_s = S^{-1}A$, and $S=\Diag(s)$.
\end{definition}

\begin{definition}[Binary hypothesis testing for leverage score model]
Let $A,B\in \R^{n\times d}$ be two matrices.
Let $P_0 = \Leverage_A$, $P_1 = \Leverage_B$ be two leverage score models.
Let $P$ be the leverage score model which is either $P_0$ or $P_1$.
In each query, we can feed $s\in (\R\backslash \{0\})^n$ into $P$, and retrieve a sample $y\in [n]$ from $P(s)$.
The goal is to determine whether the model $P$ is $P_0$ or $P_1$ in as few samples as possible.
We say an algorithm successfully distinguishes $P_0$ and $P_1$, if the correctness probability is at least $2/3$ under both hypotheses.
\end{definition}

Similar to the softmax model case, if we do not put any restrictions on $s$, then there will be certain weird behavior. For example, if we take $n=2$, $d=1$, $A = \begin{bmatrix}1 \\ 0\end{bmatrix}$ and $B = \begin{bmatrix} 1 \\ \epsilon \end{bmatrix}$ for some small $\epsilon>0$. Because $A$ and $B$ are close to each other, we should expect it to be difficult to distinguish $\Leverage_A$ and $\Leverage_B$. However, if we allow any $s\in (\R\backslash \{0\})^n$ as input, then we can take $s = \begin{bmatrix} 1 & \delta \end{bmatrix}$ for some very small $\delta>0$. In this way, we can verify that $\Leverage_A(s)$ has all mass on $1\in [n]$, while $\Leverage_B(s)$ has almost all mass on $2\in [n]$. So we can distinguish the two models using only one query. To avoid such cases we put additional constraints on $s$.
\begin{definition}[Constraint for leverage score model] \label{def:leverage-constraint}
We assume that input $s\in (\R\backslash \{0\})^d$ should satisfy the constraint such that $c \le s_i^2 \le C$ for some given constants $0<c<C$.
\end{definition}

\section{Softmax Model} \label{sec:softmax}

In Section~\ref{sec:softmax:general}, we state our general result of softmax model. In Section~\ref{sec:softmax:lower}, we provide show how to prove the lower bound. In Section~\ref{sec:softmax:upper}, we explain how to prove the upper bound.

\subsection{General Result}\label{sec:softmax:general}
We first prove a general result that relates the binary hypothesis testing problem with Hellinger distance, and the proof is deferred to Appendix~\ref{sec:missing-proof:softmax-general}.
\begin{theorem} \label{thm:softmax-general}
Let $A,B\in \R^{n\times d}$ be two matrices. Consider the binary hypothesis testing problem of distinguishing $\Softmax_A$ and $\Softmax_B$ using energy-constrained queries (Definition~\ref{def:softmax-constraint}).
Define 
\begin{align*}
    \delta = \sup_{x: \|x\|_2\le E} H(\Softmax_A(x), \Softmax_B(x)).
\end{align*}
Then the sample complexity of the binary hypothesis testing problem is $\Theta(\delta^{-2})$.
That is, there is an algorithm that successfully solves the problem using $O(\delta^{-2})$ energy-constrained queries, and any algorithm that successfully solves the problem uses $\Omega(\delta^{-2})$ energy-constrained queries.
\end{theorem}

\subsection{Lower Bound}\label{sec:softmax:lower}
Now, we prove the following lower bound for binary hypothesis testing for softmax models.
\begin{theorem}[Lower bound] \label{thm:softmax-testing-lower-bound}
If two softmax models (Definition~\ref{def:softmax_model})
with parameters $A \in \R^{n \times d}$ and $B \in \R^{n \times d}$ satisfy 
\begin{align*}
    \|A-B\|_{2\to \infty} \le \epsilon,
\end{align*}
which is 
\begin{align*}
    \max_{j\in [n]} \| A_{j,*} - B_{j,*} \|_2 \leq \epsilon,
\end{align*}
then any algorithm with energy constraint $E$ that distinguishes the two models with success probability $\ge \frac 23$ uses at least $\Omega(\epsilon^{-2} E^{-2})$ samples.
\end{theorem}

Before giving the proof of Theorem~\ref{thm:softmax-testing-lower-bound}, we state a lemma, and the proof is deferred to Appendix~\ref{sec:missing-proof:softmax-lower-bound}.
\begin{lemma}\label{lem:soft_max:lower_bound:infty_norm}
Let $a,b\in \R^n$ be such that $\|a-b\|_\infty \le \epsilon$.
Let $P$ be the distribution on $[n]$ with $p_i = \exp(a_i)/\langle \exp(a), {\bf 1}_n\rangle$.
Let $Q$ be the distribution on $[n]$ with $q_i = \exp(b_i)/\langle \exp(b), {\bf 1}_n\rangle$.
Then
\begin{align*}
H^2(P, Q) = O(\epsilon^2), ~~~~~
\TV(P, Q) = O(\epsilon).
\end{align*}
\end{lemma} 
Next, we prove an application of Lemma~\ref{lem:soft_max:lower_bound:infty_norm}.
\begin{corollary} \label{cor:softmax-h2-upper}
If matrices $A \in \R^{n \times d}, B \in \R^{n \times d}$ satisfy $\max_{j\in [n]} \| A_{j,*} - B_{j,*} \|_2 \leq \epsilon$, then for any $x\in \R^d$, the distributions $P=\Softmax_A(x)$ and $Q=\Softmax_B(x)$ satisfy
\begin{align*}
    H^2(P, Q) = O(\epsilon^2 \|x\|_2^2),~~~~~
    \TV(P, Q) = O(\epsilon \|x\|_2).
\end{align*}
\end{corollary}
\begin{proof}
For any $x\in \R^n$, we have
\begin{align*}
\|Ax-Bx\|_\infty = & ~ \max_{j\in [n]} |A_{j,*} x-B_{j,*} x| \\
\leq & ~ \max_{j\in [n]} \| A_{j,*} - B_{j,*} \|_2 \|x\|_2 \\
\leq & ~ \epsilon \|x\|_2.
\end{align*}
The result then follows from Lemma~\ref{lem:soft_max:lower_bound:infty_norm}.
\end{proof}

Now, we're ready to finish the proof of Theorem~\ref{thm:softmax-testing-lower-bound}.
\begin{proof}[Proof of Theorem~\ref{thm:softmax-testing-lower-bound}]
By Corollary~\ref{cor:softmax-h2-upper}, we have $H^2(\Softmax_A(x), \Softmax_B(x)) = O(\epsilon^2 E^2)$ for any $\|x\|_2\leq E$.
Therefore $\delta$ in the statement of Theorem~\ref{thm:softmax-general} satisfies $\delta^2 = O(\epsilon^2 E^2)$.
Applying Theorem~\ref{thm:softmax-general} we finish the proof.
\end{proof}

\subsection{Upper Bound}\label{sec:softmax:upper}
In the previous section, we established an $\Omega(\epsilon^{-2})$ lower bound for solving the hypothesis testing problem for the softmax model.
The upper bound is more subtle. Let us discuss a few difficulties in establishing the upper bound.
Let $A,B\in \R^{n\times d}$ be parameters of the softmax models, $x\in \R^d$ be the input vector, $P=\Softmax_A(x)=(p_1,\ldots,p_n)$, $Q=\Softmax_B(x)=(q_1,\ldots,q_n)$.
First, two different matrices $A$ and $B$ could give rise to the same softmax model. If $B = A + {\bf 1}_n^\top w$ for some $w\in \R^d$, then for any $x\in \R^d$, we have
\begin{align*}
q_i = & ~ \frac{\exp(Bx)_i}{\langle \exp(Bx), {\bf 1}_n\rangle } \\
= & ~ \frac{\exp(Ax)_i \exp(w^\top x)}{\langle \exp(Ax) \exp(w^\top x), {\bf 1}_n\rangle } \\
= & ~ \frac{\exp(Ax)_i}{\langle \exp(Ax) , {\bf 1}_n\rangle } \\
= & ~ p_i
\end{align*}
for all $i\in [d]$. Therefore in this case $\Softmax_A(x)=\Softmax_B(x)$ for all $x\in \R^d$ and it is impossible to distinguish the two models.
This issue may be resolved by adding additional assumptions such as ${\bf 1}_n^\top A = {\bf 1}_n^\top B$.
A more important issue is that $A$ and $B$ may differ only in rows with very small probability weight under any input $x$.
For example, suppose $A$ is the zero matrix, and $B$ differ with $A$ only in the first row.
For any $x\in \R^d$, the distribution $\Softmax_A(x)$ is the uniform distribution on $[d]$.
If $\|B_{1,*}-A_{1,*}\|_2 = \epsilon$, then for any $x$ with $\|x\|_2\le E$, we have
\begin{align*}
    \exp(-\epsilon E) 
    \leq & ~ \frac{\exp(Bx)_1}{\exp(Ax)_1} \\
    \leq & ~ \exp(\epsilon E).
\end{align*}
A simple calculation shows that in this case,
\begin{align*}
    H^2(P, Q) = O(\epsilon^2 E^2/n).
\end{align*}
So the sample complexity of any hypothesis testing algorithm is at least $\Omega(n/(\epsilon^2 E^2))$, which grows with $n$.
This shows that the sample complexity may depend on $n$.
Nevertheless, using Theorem~\ref{thm:softmax-general},
we show a local upper bound, which says that for fixed $A$ and fixed direction $M$, there is an algorithm that distinguishes $\Softmax_A$ and $\Softmax_{A+\epsilon M}$ using $O(\epsilon^{-2})$ queries, for small enough $\epsilon>0$.

\begin{theorem}\label{thm:softmax-testing-upper-bound-2}
Fix $A, M\in \R^{n\times d}$ where $\|M\|_{2\to \infty}=O(1)$.
For $\epsilon>0$, define $B_\epsilon = A + \epsilon M$. 
We consider the binary hypothesis testing problem with $\Softmax_A$ and $\Softmax_{B_\epsilon}$, for small $\epsilon$.
Let $\nu = \sup_{x: \|x\|_2\le E} \Var_{\Softmax_A(x)}(M x)$. Then for $\epsilon>0$ small enough, there is an algorithm that uses $O(\epsilon^{-2} \nu^{-1})$ energy-constrained queries and distinguishes between $\Softmax_A$ and $\Softmax_{B_\epsilon}$.
\end{theorem}
Proof of Theorem~\ref{thm:softmax-testing-upper-bound-2} is deferred to Appendix~\ref{sec:missing-proof:softmax-upper-bound-2}.
From Theorem~\ref{thm:softmax-testing-upper-bound-2} we see that it is an interesting problem to bound $\nu = \sup_{x: \|x\|_2\le E} \Var_{\Softmax_A(x)}(M x)$ for fixed $A, M\in \R^{n\times d}$. For different $A$ and $M$ the value of $\nu$ can be quite different. For example, if $A$ is the all zero matrix and $M$ is zero except for row $1$ (and $\|M\|_{2\to \infty}=O(1)$), then $\nu = O(E^2/n)$ for any $\|x\|_2 \le E$.
On the other hand, if $A$ is the zero matrix, and the first column $M$ are i.i.d.~Gaussian $\N(0,\Theta(1))$, then with high probability, $\nu = \Omega(E^2)$ for $x = (E, 0,\ldots,0)$.
We remark that Theorem~\ref{thm:softmax-testing-upper-bound-2} is in fact tight. We have a matching lower bound.
\begin{theorem} \label{thm:softmax-testing-lower-bound-2}
Under the same setting as Theorem~\ref{thm:softmax-testing-upper-bound-2} and let $\nu$ be defined as Theorem~\ref{thm:softmax-testing-upper-bound-2}, for sufficient small $\epsilon>0$, any algorithm that distinguishes between $\Softmax_A$ and $\Softmax_{B_\epsilon}$ must use $\Omega(\epsilon^{-2} \nu^{-1})$ energy-constrained queries.
\end{theorem}
\begin{proof}
It follows from combining the proof of Theorem~\ref{thm:softmax-testing-upper-bound-2} and Theorem~\ref{thm:softmax-general}.
\end{proof}
\section{Leverage Score Model} \label{sec:leverage}

In Section~\ref{sec:leverage:general}, we state our general result of softmax model. In Section~\ref{sec:leverage:lower}, we provide show how to prove the lower bound. In Section~\ref{sec:leverage:upper}, we explain how to prove the upper bound.

\subsection{General Result}\label{sec:leverage:general}
We first prove a general result which is the leverage score version of Theorem~\ref{thm:softmax-general}.
\begin{theorem} \label{thm:leverage-general}
Let $A,B\in \R^{n\times d}$ be two matrices. Consider the binary hypothesis testing problem of distinguishing $\Leverage_A$ and $\Leverage_B$ using constrained queries (Definition~\ref{def:leverage-constraint}).
Define 
\begin{align*}
    \delta = \sup_{s: c \le s_i^2\le C \forall i} H(\Leverage_A(s), \Leverage_B(s)).
\end{align*}
Then the sample complexity of the binary hypothesis testing problem is $\Theta(\delta^{-2})$.
That is, there is an algorithm that successfully solves the problem using $O(\delta^{-2})$ energy-constrained queries, and any algorithm that successfully solves the problem uses $\Omega(\delta^{-2})$ energy-constrained queries.
\end{theorem}
\begin{proof}
The proof is similar to Theorem~\ref{thm:softmax-general} and omitted.
\end{proof}

\subsection{Lower Bound}\label{sec:leverage:lower}

The goal of this section is to prove the following lower bound for binary hypothesis testing for leverage score models.
\begin{theorem}\label{thm:leverage-testing-lower-bound}
Consider two leverage score model $\Leverage_A$ and $\Leverage_B$.
Assume that there exists $\delta>0$ such that $A^\top A \succeq \delta I$.
If 
\begin{align*}
\sum_{i\in [n]} \|B_{i,*}^\top B_{i,*} - A_{i,*}^\top A_{i,*}\|_{\op} \le \epsilon
\end{align*}
(where $\|\cdot\|_{\op}$ denotes the $2$-to-$2$ operator norm), then any algorithm that solves the binary hypothesis testing problem takes at least $\Omega(c \delta / (C \epsilon))$ constrained queries.
\end{theorem}
\begin{proof}
Let 
\begin{align*}
    P=\Leverage_A(s)=(p_1,\ldots,p_n)
\end{align*}
and
\begin{align*}
    Q=\Leverage_B(s)=(q_1,\ldots,q_n).
\end{align*}
By Theorem~\ref{thm:leverage-general}, it suffices to prove that $H^2(P, Q) = O(\epsilon C /(c \delta))$.
We first consider the case where $A$ and $B$ differ in exactly one row $i$.
Fix $s\in \R^d$ with $c\le s_j\le C$ for all $j\in [n]$.
Let $A_s = S^{-1} A$ and $B_s = S^{-1} B$, where $S=\Diag(s)$.

Because $A^\top A \succeq \delta I$, we have 
\begin{align*}
    A_s^\top A_s \succeq (\delta/C) \cdot I.
\end{align*}
Because $\|B_{i,*}^\top B_{i,*} - A_{i,*}^\top A_{i,*}\|_{\op} \le \epsilon$, we have
\begin{align*}
-\epsilon_i C/\delta A_s^\top A_s \preceq B_{i,*}^\top B_{i,*} - A_{i,*}^\top A_{i,*} \preceq \epsilon_i C/\delta A_s^\top A_s.
\end{align*}
Recall that $A$ and $B$ differ in exactly one row $i$. Therefore 
\begin{align}\label{eq:bound_for_B_s}
(1-\frac{\epsilon C}{c\delta}) A_s^\top A_s \preceq B_s^\top B_s \preceq (1+\frac{\epsilon C}{c\delta}) A_s^\top A_s.
\end{align}

For $j\ne i$, we have
\begin{align}\label{eq:lower_bound_q_j}
q_j &= s_j^{-2} B_{j,*} (B_s^\top B_s)^{-1} (B^\top)_{*,j} /d \notag \\
& = \tr [ s_j^{-2} (B^\top)_{*,j} B_{j,*} (B_s^\top B_s)^{-1} ] /d \notag \\
&= (1 \pm O(\epsilon C/(c \delta))) \tr [ s_j^{-2} A_{j,*}^\top A_{j,*} (A_s^\top A_s)^{-1} ] /d \notag \\
&= (1 \pm O(\epsilon C/(c \delta))) p_j,
\end{align}
where the first step is by definition of the leverage score model, the second step is by property of trace, the third step is Eq.~\eqref{eq:bound_for_B_s}, the fourth step is by definition of the leverage score model.

{\bf Upper bound for $\TV$.}
For the TV distance, we have
\begin{align*}
    \TV(P,Q) = & ~ \frac 12 \sum_{j=1}^n  |p_j-q_j| \\
    \leq & ~ \sum_{j\neq i}  |p_j-q_j| \\
    \leq & ~ \sum_{j\neq i}  O(\epsilon C/(c \delta)) p_i \\
    \leq & ~ O(\epsilon C/(c \delta)).
\end{align*}
where the first step is by definition of TV distance, and the third step is by Eq.~\eqref{eq:lower_bound_q_j}.
Therefore $\TV(P,Q) \le O(\epsilon C/(c \delta))$.

{\bf Upper bound for $H^2(P,Q)$.}
Using 
\begin{align*}
    H^2(P,Q)\le \TV(P,Q),
\end{align*}
we also get 
\begin{align*}
    H^2(P,Q) \le O(\epsilon C/(c \delta)).
\end{align*}

Now we have established the result when $A$ and $B$ differ in exactly one row. Let us now consider general case.
If $\epsilon \ge 0.1 \delta$, then $c \delta / (C \epsilon)=O(1)$ and there is nothing to prove. In the following, assume that $\epsilon \le 0.1 \delta$.
For $0\le k\le n$, define $B^k\in \R^{n\times d}$ be the matrix with $B^k_{i,*}=B_{i,*}$ for $i\le k$ and $B^k_{i,*}=A_{i,*}$ for $i\ge k$.
Then $B^0=A$, $B^n=B$, and $B^k$ and $B^{k+1}$ differ exactly in one row.
Let $\epsilon_i = \|B_{i,*}^\top B_{i,*} - A_{i,*}^\top A_{i,*}\|_{\op}$.

Then by the above discussion, we have
\begin{align*}
\TV(\Leverage_{B^k}(s), \Leverage_{B^{k+1}}(s)) 
= & ~ O(\epsilon_k C / (c \delta))
\end{align*}
for all $0\le k\le n-1$.
By metric property of $\TV$, we have 
\begin{align*}
& ~ \TV(P, Q) \\
\le & ~ \sum_{0\le k\le n-1}\TV(\Leverage_{B^k}(s), \Leverage_{B^{k+1}}(s)) \\
= & ~ \sum_{0\le k\le n-1} O(\epsilon_i C / (c \delta) ) \\
= & ~ O(\epsilon C /(c \delta)).
\end{align*}
Using $H^2(P, Q) \le \TV(P, Q)$ we also get $H^2(P, Q) = O(\epsilon C /(c \delta))$. This finishes the proof.
\end{proof}

In Theorem~\ref{thm:leverage-testing-lower-bound}, the bound has linear dependence in $\epsilon^{-1}$. An interesting question is the improve the bound to quadratic dependence $\epsilon^{-2}$.

\subsection{Upper Bound}\label{sec:leverage:upper}
Let $A,B\in \R^{n\times d}$ be parameters of the leverage score models, $s\in \R^n$ be the input vector, $P = \Leverage_A(s) = (p_1,\ldots,p_n)$, $Q = \Leverage_B(s) = (q_1,\ldots, q_n)$.
For the upper bounds of the leverage score model, we run into similar difficulties as for the softmax model. 
Firstly, different matrices $A$ and $B$ could give rise to the same leverage score model. If $B = A R$ for some invertible matrix $R\in \R^{d\times d}$, then we have
\begin{align*}
q_i = & ~ (B_s (B_s^\top B_s)^{-1} B_s^\top)_{i,i}/d\\
= & ~ (A_s R (R^\top A_s^\top A_s R)^{-1} R^\top A_s^\top)_{i,i}/d \\
= & ~ (A_s (A_s^\top A_s)^{-1} A_s^\top)_{i,i}/d \\
= & ~ p_i.
\end{align*}
Then $\Leverage_A(s)=\Leverage_B(s)$ for all $s\in (\R\backslash \{0\})^n$ and it is impossible to distinguish the two models.
Furthermore, there exist scenarios where $A$ and $B$ differ only in rows with very small probability weight under any input $s$.

We now give an example where 
\begin{align*}
    \|A_{1,*}^\top A_{1,*}-B_{1,*}^\top B_{1,*}\|=\Omega(1)
\end{align*}
but 
\begin{align*}
\TV(\Leverage_A(s),\Leverage_B(s))=O(1/n)
\end{align*}
for any $s$ satisfying $c \le s_i^2 \le C$ for all $i\in [n]$.

Suppose 
$
    A = \begin{bmatrix}I_d & e_1 &\cdots& e_1\end{bmatrix}^\top,
$
that is, the first $d$ rows of $A$ is equal to $I_d$, and all remaining rows are equal to $e_1^\top = (1,0,\ldots,0)$.

Then for $s$ satisfying $c \le s_i^2 \le C$ for all $i\in [n]$, the distribution $P=\Leverage_A(s)$ has probability mass $O(1/n)$ on every element $i\in \{1,d+1,d+2\ldots,n\}$ (hiding constants depending on $c$ and $C$).

Now suppose $B$ differs with $A$ only in the first entry $(1,1)$, and 
\begin{align*}
    B_{1,1} = A_{1,1} +\Theta(1).
\end{align*}
Then for fixed $s$, $q_j=p_j$ for $j\in \{2,\ldots,d\}$, $q_1\ge p_1$, and $q_j\le p_j$ for $j\in \{d+1,\ldots,n\}$.
So 
\begin{align*}
    H^2(P, Q) \leq & ~ \TV(P, Q) \\
    = & ~ q_1-p_1 \\
    = & ~ \Theta(1/n).
\end{align*}
This shows that the sample complexity may depend on $n$.
After discussing the difficulties in establishing an upper bound, we now show a local upper bound, which says for fixed $A$ and fixed direction $M$, there is an algorithm that distinguishes $\Leverage_A$ and $\Leverage_{A+\epsilon M}$ using $O(\epsilon^{-2})$ queries, for small enough $\epsilon>0$.

\begin{theorem}\label{thm:leverage-testing-upper-bound-2}
Fix $A,M\in \R^{n\times d}$ where $\|M\|_{2\to \infty} = O(1)$.
For $\epsilon>0$, define $B_\epsilon = A + \epsilon M$. We consider the binary hypothesis testing problem with $\Leverage_A$ and $\Leverage_{B_\epsilon}$, for small $\epsilon$.
Let $\nu = \sup_s \Var_{\Leverage_A(s)}(w_s)$ where 
\begin{align*}
w_s =  \frac{ \diag((I-A_s(A_s^\top A_s)^{-1} A_s^\top)(M_s (A_s^\top A_s)^{-1} A_s^\top))}{ \diag(A_s(A_s^\top A_s)^{-1} A_s^\top) }
\end{align*}
where the division between vectors is entrywise division. Then for $\epsilon>0$ small enough, there is an algorithm that uses $O(\epsilon^{-2} \nu^{-1})$ queries and distinguishes between $\Leverage_A$ and $\Leverage_{B_\epsilon}$.
\end{theorem}
Proof of Theorem~\ref{thm:leverage-testing-upper-bound-2} is deferred to Appendix~\ref{sec:missing-proof:leverage-upper-bound-2}.
Similarly to the softmax model case, Theorem~\ref{thm:leverage-testing-upper-bound-2} is also tight.
\begin{theorem} \label{thm:leverage-testing-lower-bound-2}
Work under the same setting as Theorem~\ref{thm:leverage-testing-upper-bound-2}.
For $\epsilon>0$ small enough, any algorithm that distinguishes between $\Softmax_A$ and $\Softmax_{B_\epsilon}$ must use $\Omega(\epsilon^{-2} \nu^{-1})$ energy-constrained queries.
\end{theorem}
\begin{proof}
The proof is by combining the proof of Theorem~\ref{thm:leverage-testing-upper-bound-2} and Theorem~\ref{thm:leverage-general}. We omit the details.
\end{proof}
\section{Discussion}\label{sec:discussion}

\paragraph{Practical Implications.} Softmax and leverage score distributions arise naturally in machine learning and numerical linear algebra~\cite{ls14,lsw15,cls19,s19,jkl+20,swz19,lls+20}. Softmax distributions have a clear connection to LLMs~\cite{wwz+22,ll21,ddh+21,byks22,hbkg23,xqp+22}, and leverage score distribution serves as a more general and complex case. We study the distinguishability of models (softmax and leverage score based) through the lens of
binary hypothesis testing, establishing tight sample complexity bounds. These results directly address the challenge of determining how much information (or how many queries) is
needed to tell apart closely related models, a theoretical formulation aligned with understanding model ``abilities'' via limited parameter access. Moreover, our framework sets a path toward identifying distinguishable components of large models. For instance, showing that certain parameters contribute more significantly to distinguishability (via Hellinger distance
bounds) ofers insight into what might constitute an “ability region” within a model, aligning with the introductory motivation.

\paragraph{Technical Novelty.} The core novelty of our proofs lies in adapting classical binary hypothesis testing~\cite{pajl23,pjl24}, typically studied in the context of generic distributions, to the structured,
parameterized families of distributions induced by softmax and leverage score models. Unlike arbitrary distributions, these models produce distributions that are nonlinear functions
of the input and matrix parameters, which poses unique analytical challenges.

For the softmax model (Section~\ref{sec:softmax}), one key difficulty is that different parameter matrices and can induce indistinguishable distributions due to invariance under certain transformations (e.g.,
row shifts). To handle this, we introduce structural constraints and prove that the Hellinger distance~\cite{polyanskiy2023information} between softmax outputs under constrained inputs governs
the sample complexity. This leads to a tight upper and lower bound framework via careful analysis of the sensitivity of softmax distributions to perturbations in the parameter matrix.

For the leverage score model (Section~\ref{sec:leverage}), the challenge is even greater due to the nonlinear matrix expressions involved, including matrix inversion and normalization, and the fact that the input
sss is a vector that rescales rows of the matrix. We overcome this by establishing operator norm bounds and using perturbation theory to relate changes in the parameter matrix to
changes in the output distribution. Our proof carefully propagates these changes through the matrix expressions and yields tight dependence on both the model difference and input constraints.
\section{Conclusion and Future Directions}\label{sec:conclusion}
Widely applied across various domains, softmax and leverage scores play crucial roles in machine learning and linear algebra. This study delves into the testing problem aimed at distinguishing between different models of softmax and leverage score distributions, each parameterized by distinct matrices. We establish bounds on the number of samples within the defined testing problem.
With the rapidly escalating computational costs in current machine learning research, our work holds the potential to offer valuable insights and guidance for distinguishing between the distributions of different models.
We discuss a few possible directions for further research.
In Theorem~\ref{thm:softmax-testing-upper-bound-2} and Theorem~\ref{thm:leverage-testing-upper-bound-2}, we determine the local sample complexity of the binary hypothesis testing problems for softmax models and leverage score models. In particular, the sample complexity is $\Theta(\epsilon^{-2} \nu)$, where $\nu$ is a certain function depending on $A$ and $M$ (where $B = A + \epsilon M$). The form of $\nu$ is an optimization problem over the space of possible inputs. An interesting question is to provide bounds on the quantity $\nu$, or to provide computation-efficient algorithms for determining the value of $\nu$ of finding the optimal input ($x$ for softmax, $s$ for leverage score). This will lead to computation-efficient algorithms for solving the binary hypothesis testing problem in practice.

In this paper, we focused on the binary hypothesis testing problem, where the goal is to distinguish two models with different parameters. There are other hypothesis testing problems that are of interest both in theory and practice. For example, in the goodness-of-fit problem, the goal is to determine whether an unknown model is equal to or far away from a given model. In the two-sample testing problem, the goal is to determine whether two unknown models are the same or far away from each other. These problems have potential practical applications and we leave them as an interesting future direction.

\section*{Acknowledgements}
The author would like to thank the anonymous reviewer of ICML 2025 for their highly insightful suggestions.

\section*{Impact Statement}

This paper presents work whose goal is to advance the field of Machine Learning. There are many potential societal consequences of our work, none of which we feel must be specifically highlighted here.

\ifdefined\isarxiv

\else
\bibliography{ref}
\bibliographystyle{icml2025}
\fi

\newpage
\onecolumn
\appendix

\begin{center}
    \textbf{\LARGE Appendix }
\end{center}

{\bf Roadmap.}
In Section~\ref{sec:missing-proof}, we provide all proofs which are missing from the main text.
In Section~\ref{sec:more_related}, we present more related work.

\section{Missing Proofs} \label{sec:missing-proof}
This section provide missing proofs from the main text.
In Section~\ref{sec:missing-proof:softmax-general}, we show the proof of general result for softmax model.
In Section~\ref{sec:missing-proof:softmax-lower-bound}, we demonstrate the proof of the lower bound for softmax model.
In Section~\ref{sec:missing-proof:softmax-upper-bound-2}, we prove the local upper bound for softmax model.
In Section~\ref{sec:missing-proof:leverage-upper-bound-2}, we present the proof for local upper bound for leverage score model.

\subsection{General Result for Softmax Model} \label{sec:missing-proof:softmax-general}
\begin{proof}[Proof of Theorem~\ref{thm:softmax-general}]
{\bf Lower bound.}
If $\delta \ge 0.1$ then there is nothing to prove. In the following assume that $\delta < 0.1$.
Suppose that there is an algorithm that successfully solves the binary hypothesis testing problem.
Suppose it makes queries $x_1,\ldots, x_m\in \R^d$ where $x_i$ may depend on previous query results.
Let $Y_1,\ldots,Y_m\in [n]$ denote the query results.
Let $P_{Y^m}$ and $Q_{Y^m}$ denote the distribution of $Y^m$ under $P$ and $Q$, respectively.
By definition of $\delta$, we have
\begin{align*}
H^2(P_{Y_k|Y^{k-1}},Q_{Y_k|Y^{k-1}}) \le \delta^2.
\end{align*}
for any $k\in [m]$ and $Y^{k-1}$.
Then
\begin{align*}
& ~ 1-H^2(P_{Y^m}, Q_{Y^m}) \\
= & ~ 
\int \sqrt{P_{y^m} Q_{y^m}} \d y^m \\
= & ~ \int \sqrt{P_{y^{m-1}} Q_{y^{m-1}}}  \big(\int \sqrt{P_{y_m|y^{m-1}} Q_{y_m|y^{m-1}}} dy_m \big) \d y^{m-1} \\
\ge & ~ \int \sqrt{P_{y^{m-1}} Q_{y^{m-1}}} (1-\delta^2) \d y^{m-1}.
\end{align*}

Repeating this computation, in the end we get
\begin{align*}
1-H^2(P_{Y^m}, Q_{Y^m}) \ge (1-\delta^2)^m.
\end{align*}
Because $\delta \le 0.1$, we have $1-\delta^2 \ge \exp(-2\delta^2)$.
If $m \le 0.01 \delta^{-2}$,
then 
\begin{align*}
    1-H^2(P_{Y^m}, Q_{Y^m}) 
    \ge & ~ \exp(-2\delta^2 m) \\
    \ge & ~ \exp(-0.02) \\
    > & ~ 0.98,
\end{align*}
and 
\begin{align*}
    H^2(P_{Y^m}, Q_{Y^m}) \le 0.02.
\end{align*}
This implies 
\begin{align*}
\TV(P_{Y^m}, Q_{Y^m}) 
\le  \sqrt 2 H(P_{Y^m}, Q_{Y^m}) 
\le  0.2,
\end{align*}
which implies the success rate for binary hypothesis testing cannot be $\ge \frac 23$.

In conclusion, any algorithm that successfully solves the hypothesis testing problem need to use $\Omega(\delta^{-2})$ queries.

{\bf Upper bound.}
Take $x\in \R^d$ such that $\|x\|_2\le E$ and $\delta = H(\Softmax_A(x), \Softmax_B(x))$.
By Lemma~\ref{lem:binary-hypo-test-classic}, using $O(\delta^{-2})$ samples we can distinguish $\Softmax_A(x)$ and $\Softmax_B(x)$.
Therefore we can distinguish $\Softmax_A$ and $\Softmax_B$ in $O(\delta^{-2})$ queries by repeatedly querying $x$.
\end{proof}

\subsection{Lower Bound for Softmax Model} \label{sec:missing-proof:softmax-lower-bound}
Before giving the proof of Lemma~\ref{lem:soft_max:lower_bound:infty_norm}, we prove a weaker version of the lemma.
\begin{lemma}\label{lem:soft_max:lower_bound:0_norm}
    Let $a,b\in \R^n$. Suppose there exists an $\epsilon\ge 0$ such that for every $i\in [n]$, $b_i-a_i\in \{0, \epsilon\}$.
    Let $P$ be the distribution on $[n]$ with $p_i = \exp(a_i)/\langle \exp(a), {\bf 1}_n\rangle$.
    Let $Q$ be the distribution on $[n]$ with $q_i = \exp(b_i)/\langle \exp(b), {\bf 1}_n\rangle$.
    Then
    \begin{align*}
    H^2(P, Q) & = \frac{(1-\exp(\epsilon/4))^2}{1+\exp(\epsilon/2)} = O(\epsilon^2), \\
    \TV(P, Q) & = \tanh(\epsilon/4) = O(\epsilon).
    \end{align*}
\end{lemma}
\begin{proof}

Assume that $a$ and $b$ differ in $m$ coordinates.
By permuting the coordinates, WLOG assume that $b_i=a_i+\epsilon$ for $1\le i\le m$ and $b_i=a_i$ for $m+1\le i\le n$.

Write 
\begin{align*}
s = \sum_{i=1}^{m} \exp(a_i)
\end{align*}
and 
\begin{align*}
t = \sum_{ i = m+1 }^{ n } \exp(a_i).
\end{align*}

Then
\begin{align*}
H^2(P, Q) 
= & ~ 1 - \sum_{i\in [n]} \sqrt{p_i q_i} \\
= & ~ 
1 - \frac{s \exp(\epsilon/2) + t}{\sqrt{(s+t) (s \exp(\epsilon)+t)}}.
\end{align*}
For fixed $t$ and $\epsilon$, the above is maximized at 
\begin{align*}
    s = t \exp(-\epsilon/2).
\end{align*}

Plugging in the above $s$, we get
\begin{align*}
H^2(P, Q) 
\le & ~ 1 - \frac{2}{\sqrt{(\exp(-\epsilon/2)+1)(\exp(\epsilon/2)+1)}} \\
= & ~ \frac{(1-\exp(\epsilon/4))^2}{1+\exp(\epsilon/2)}.
\end{align*}

For $\TV$, we have 
\begin{align*}
\TV(P, Q) 
= & ~ \sum_{m+1\le i\le n} (q_i - p_i) \\
= & ~ \frac{t}{s+t} - \frac{t}{s \exp(\epsilon) + t}.
\end{align*}
For fixed $t$ and $\epsilon$ the above is maximized at $s = t \exp(-\epsilon/2)$.
Plugging in this $s$, we get
\begin{align*}
\TV(P, Q) \le \tanh(\epsilon/4).
\end{align*}
\end{proof}

\begin{proof}[Proof of Lemma~\ref{lem:soft_max:lower_bound:infty_norm}]
We first prove the case where $b_i\ge a_i$ for all $i\in [n]$.
Define $\epsilon_i=b_i-a_i$ for all $i\in [n]$.
By permuting the coordinates, WLOG assume that $\epsilon_1 \le \cdots \le \epsilon_n$. Specially, define $\epsilon_0=0$.
For $0\le k\le n$, let $b^k\in \R^n$ denote the vector where $b^k_i = a_i + \min\{\epsilon_i,\epsilon_k\}$ for all $i\in [k]$.
Then we can see that $b^0=a$ and $b^n=b$, and for every $0\le k\le n-1$, the pair $(b^k, b^{k+1})$ satisfies the assumption in Lemma~\ref{lem:soft_max:lower_bound:0_norm}.
For $0\le k\le n$, let $P^k$ denote the softmax distribution corresponding to $b^k$.
By Lemma~\ref{lem:soft_max:lower_bound:0_norm}, for every $0\le k\le n-1$, we have
\begin{align*}
H(P^k,P^{k+1}) &= O(\epsilon_{k+1}-\epsilon_k),\\
\TV(P^k,P^{k+1}) &= O(\epsilon_{k+1}-\epsilon_k).
\end{align*}
Because Hellinger distance and TV distance are both metrics, we have
\begin{align*}
H(P,Q) 
= & ~ H(P^0,P^n) \\
\le & ~ \sum_{ k = 0}^{n-1} H(P^k,P^{k+1}) \\
= & ~ O(\epsilon), 
\end{align*}
and 
\begin{align*}
\TV(P,Q)
= & ~ \TV(P^0,P^n) \\
\le & ~ \sum_{k=0}^{n-1} \TV(P^k,P^{k+1}) \\
= & ~ O(\epsilon).
\end{align*}
This finishes the proof of the result when $b_i\ge a_i$ for all $i\in [n]$.

Now let us consider the general case.
Let $c\in \R^n$ be defined as $c_i = \max\{a_i,b_i\}$ for all $i\in [n]$.
Then 
\begin{align*}
  \max\{\|a-c\|_\infty,\|c-b\|_\infty\} \le \|a-b\|_\infty \le \epsilon.
\end{align*}
Let $R$ be the softmax distribution corresponding to $c$.
By our previous discussion, we have
\begin{align*}
H(P,R),H(R,Q), \TV(P,R),\TV(R,Q) = O(\epsilon).
\end{align*}
By metric property of Hellinger distance and TV distance, we get
\begin{align*}
H(P,Q), H(P,Q) = O(\epsilon)
\end{align*}
as desired.

\end{proof}

\subsection{Local Upper Bound for Softmax Model} \label{sec:missing-proof:softmax-upper-bound-2}
\begin{proof}[Proof of Theorem~\ref{thm:softmax-testing-upper-bound-2}]
We take an $x$ satisfying $\|x\|_2\le E$ that maximizes $\Var_{\Softmax_A(x)}(M x)$ and repeatedly query $x$.
We would like to apply Theorem~\ref{thm:softmax-general}.
To do that, we need to show that
\begin{align*}
H^2(\Softmax_A(x), \Softmax_{B_\epsilon}(x)) = \Omega(\epsilon^2 \nu).
\end{align*}
Let $P = \Softmax_A(x) = (p_1,\ldots,p_n)$, $Q_\epsilon = \Softmax_{B_\epsilon}(x) = (q_{\epsilon,1},\ldots,q_{\epsilon,n})$.
Write $Z_A = \langle \exp(A x), {\bf 1}_n\rangle $, $Z_{B_\epsilon} = \langle \exp(B_\epsilon x), {\bf 1}_n\rangle $. 

Then, it follows that
\begin{align}\label{eq:Z_B}
Z_B &= \sum_{j\in [n]} \exp(A x)_j \exp(\epsilon (M x)_j)\notag\\
&= \sum_{j\in [n]} \exp(A x)_j + \sum_{j\in [n]} \exp(A x)_j (\exp(\epsilon (M x)_j)-1) \notag\\
&= \sum_{j\in [n]} \exp(A x)_j + \sum_{j\in [n]} \exp(A x)_j (\epsilon (M x)_j + O(\epsilon^2) ) \notag\\
&= Z_A (1 + \epsilon\langle p, M x\rangle + O(\epsilon^2)).
\end{align}
where the initial step is because of $B = A + \epsilon M$, the second step is a result of simple algebra, the third step is a consequence of the Taylor expansion of $\exp(\cdot)$, assuming $\epsilon$ is sufficiently small and the fourth step is the result of the definition of $Z_A$ and involves the consolidation of addition, introducing the common term $Z_A$.

Then
\begin{align}\label{eq:q_eps_i}
q_{\epsilon,i} &= \frac{\exp(B_\epsilon x)_i}{ Z_B } \notag\\
& = \frac{\exp(A x)_i \exp(\epsilon M x)_i}{ Z_A (1 + \epsilon\langle p, M x\rangle + O(\epsilon^2))} \notag\\
& = p_i (1 + \epsilon ((M x)_i - \langle p, M x\rangle) + O(\epsilon^2)).
\end{align}
where the initial step is because of the definition of $q_{\epsilon,i}$, the subsequent step is a result of Eq.\eqref{eq:Z_B}, and the third step is due to the definition of $q_i$ along with the Taylor expansion of $f(x) = 1/(1 + x)$ and $\exp(\cdot)$, considering $\epsilon$ as a sufficiently small value.

So, we have that
\begin{align*}
H^2(P,Q_\epsilon)
& = \frac 12 \sum_{i =1}^n (\sqrt{p_i}-\sqrt{q_{\epsilon,i}})^2 \\
& = \frac 12 \sum_{i = 1}^n p_i (\epsilon^2 ((M x)_i- \langle p, M x\rangle)^2 + O(\epsilon^3)) \\
& = \frac 12 \epsilon^2 \Var_{P}(M x) + O(\epsilon^3) \\
& = \frac 12 \epsilon^2 \nu + O(\epsilon^3).
\end{align*}
where the first step is the result of Definition~\ref{def:hellinger_distance}, the second step is because of Eq.\eqref{eq:q_eps_i}, the third step the result of definition of $\Var_{P}(M x)$ (See Definition~\ref{def:var}) and the forth step follows from the expression $\nu = \sup_{x: \|x\|_2\le E} \Var_{\Softmax_A(x)}(M x)$.

Applying Theorem~\ref{thm:softmax-general} we finish the proof.
\end{proof}

\subsection{Local Upper Bound for Leverage Score Model}\label{sec:missing-proof:leverage-upper-bound-2}
\begin{proof}[Proof of Theorem~\ref{thm:leverage-testing-upper-bound-2}]
We take an $s$ satisfying $c\le s_i^2\le C$ and $\forall i\in [n]$ that maximizes $\sup_s \Var_{\Leverage_A(s)}(w_s)$ and repeatedly query $s$.
We need to show that
\begin{align*}
H^2(\Leverage_A(s), \Leverage_{B_\epsilon}(s)) = \Omega(\epsilon^2 \nu).
\end{align*}
Let $P=\Leverage_A(s)=(p_1,\ldots,p_n)$, $Q_\epsilon=\Leverage_{B_\epsilon}(x)=(q_{\epsilon,1},\ldots,q_{\epsilon,n})$.
We can compute that
\begin{align*}
\frac{d}{d\epsilon} q_{\epsilon,i} = (2(I-A_s(A_s^\top A_s)^{-1} A_s^\top)(M_s (A_s^\top A_s)^{-1} A_s^\top))_{i,i}.
\end{align*}
Define $W = (I-A_s(A_s^\top A_s)^{-1} A_s^\top)(M_s (A_s^\top A_s)^{-1} A_s^\top)$.
Then
\begin{align*}
q_{\epsilon,i} = p_i + 2W_{i,i} \epsilon + O(\epsilon^2).
\end{align*}
Computing $H^2(P, Q_\epsilon)$ we get
\begin{align*}
H^2(P, Q_\epsilon) &= 
\frac 12 \sum_{i\in [n]} (\sqrt{q_{\epsilon,i}}-\sqrt{p_i})^2 \\
&=\sum_{i\in [n]} p_i \left(\frac{W_{i,i}}{p_i} \epsilon + O(\epsilon^2)\right)^2\\
&=\sum_{i\in [n]} \frac{W_{i,i} \epsilon^2}{p_i} + O(\epsilon^3)\\ 
&= \epsilon^2 \nu + O(\epsilon^3).
\end{align*}
\end{proof}

\section{More Related Work}
\label{sec:more_related}

\paragraph{Softmax Computation and Regression}
Softmax computation, a crucial element in attention computation \citep{vsp+17}, plays a pivotal role in the development of LLMs. Several studies \cite{as23, bsz23, lsz+23, dms23} delve into the efficiency of softmax computation.
To improve computational efficiency, \cite{as23} presents a quicker attention computation algorithm utilizing implicit matrices. Similarly, \cite{bsz23} utilizes lazy updates to speed up dynamic computation, while \cite{dms23} employs a randomized algorithm for similar efficiency gains. Conversely, \cite{lsz+23} utilizes an approximate Newton method that operates in nearly linear time.
\cite{gms23} centers on the convergence of overparameterized two-layer networks with exponential activation functions, whereas \cite{dls23, lsz+23} explore regression analysis within the framework of attention computation. All of these studies specifically focus on softmax-based regression problems. 

Softmax functions is also widely used in computer vision. In particular, it is widely used in different backbone models, such as Vision Transformers (ViTs)~\cite{dbk+20,zlz21,llc+21,ylpm24} and Visual Autoregressive (VAR)~\cite{tjy+24} models, and also has advanced many applications including diffusion models~\cite{px23,csy25,ssz+25}, flow matching~\cite{lcb+22,lgl22,ccl+25}, and GANs~\cite{bds19,ttn+19,lwz+22}. 
Beyond computer vision, there are also other ML applications that involve softmax computation, such as Hopfield networks~\cite{kja+23,hlsl24,hwl24,hcw+24,whl+24,whh+24,hzs+25}, graph neural networks~\cite{vcc+18,flz+21,gsj+22,chl+24_gat}, recommender systems~\cite{km18,llz+23,lzw+24}, and fairness problems~\cite{cgn+23,svwz24,cll+25}.

\ifdefined\isarxiv
\bibliographystyle{alpha}
\bibliography{ref}

\begin{thebibliography}{181}
\providecommand{\natexlab}[1]{#1}
\providecommand{\url}[1]{\texttt{#1}}
\expandafter\ifx\csname urlstyle\endcsname\relax
  \providecommand{\doi}[1]{doi: #1}\else
  \providecommand{\doi}{doi: \begingroup \urlstyle{rm}\Url}\fi

\bibitem[Alman \& Song(2023)Alman and Song]{as23}
Alman, J. and Song, Z.
\newblock Fast attention requires bounded entries.
\newblock \emph{Advances in Neural Information Processing Systems},
  36:\penalty0 63117--63135, 2023.

\bibitem[Alman \& Song(2024{\natexlab{a}})Alman and Song]{as24_iclr}
Alman, J. and Song, Z.
\newblock How to capture higher-order correlations? generalizing matrix softmax
  attention to kronecker computation.
\newblock In \emph{ICLR}, 2024{\natexlab{a}}.

\bibitem[Alman \& Song(2024{\natexlab{b}})Alman and Song]{as24_neurips}
Alman, J. and Song, Z.
\newblock The fine-grained complexity of gradient computation for training
  large language models.
\newblock In \emph{NeurIPS}, 2024{\natexlab{b}}.

\bibitem[Alman \& Song(2025{\natexlab{a}})Alman and Song]{as25_rank}
Alman, J. and Song, Z.
\newblock Only large weights (and not skip connections) can prevent the perils
  of rank collapse.
\newblock \emph{arXiv preprint arXiv:2505.16284}, 2025{\natexlab{a}}.

\bibitem[Alman \& Song(2025{\natexlab{b}})Alman and Song]{as25_rope}
Alman, J. and Song, Z.
\newblock Fast rope attention: Combining the polynomial method and fast
  {F}ourier transform.
\newblock In \emph{arXiv preprint arXiv:2505.11892}, 2025{\natexlab{b}}.

\bibitem[Arias-Castro et~al.(2018)Arias-Castro, Pelletier, and
  Saligrama]{arias2018remember}
Arias-Castro, E., Pelletier, B., and Saligrama, V.
\newblock Remember the curse of dimensionality: The case of goodness-of-fit
  testing in arbitrary dimension.
\newblock \emph{Journal of Nonparametric Statistics}, 30\penalty0 (2):\penalty0
  448--471, 2018.

\bibitem[Bang et~al.(2023)Bang, Cahyawijaya, Lee, Dai, Su, Wilie, Lovenia, Ji,
  Yu, Chung, et~al.]{bcl+23}
Bang, Y., Cahyawijaya, S., Lee, N., Dai, W., Su, D., Wilie, B., Lovenia, H.,
  Ji, Z., Yu, T., Chung, W., et~al.
\newblock A multitask, multilingual, multimodal evaluation of chatgpt on
  reasoning, hallucination, and interactivity.
\newblock In \emph{Proceedings of the 13th International Joint Conference on
  Natural Language Processing and the 3rd Conference of the Asia-Pacific
  Chapter of the Association for Computational Linguistics (Volume 1: Long
  Papers)}, pp.\  675--718, 2023.

\bibitem[Bourgain et~al.(1989)Bourgain, Lindenstrauss, and Milman]{blm89}
Bourgain, J., Lindenstrauss, J., and Milman, V.
\newblock Approximation of zonoids by zonotopes.
\newblock 1989.

\bibitem[Boutsidis \& Woodruff(2014)Boutsidis and Woodruff]{bw14}
Boutsidis, C. and Woodruff, D.~P.
\newblock Optimal cur matrix decompositions.
\newblock In \emph{Proceedings of the forty-sixth annual ACM symposium on
  Theory of computing}, pp.\  353--362, 2014.

\bibitem[Brand et~al.(2020{\natexlab{a}})Brand, Lee, Nanongkai, Peng,
  Saranurak, Sidford, Song, and Wang]{bln+20}
Brand, J. v.~d., Lee, Y.-T., Nanongkai, D., Peng, R., Saranurak, T., Sidford,
  A., Song, Z., and Wang, D.
\newblock Bipartite matching in nearly-linear time on moderately dense graphs.
\newblock In \emph{2020 IEEE 61st Annual Symposium on Foundations of Computer
  Science (FOCS)}, pp.\  919--930. IEEE, 2020{\natexlab{a}}.

\bibitem[Brand et~al.(2020{\natexlab{b}})Brand, Lee, Sidford, and Song]{blss20}
Brand, J. v.~d., Lee, Y.~T., Sidford, A., and Song, Z.
\newblock Solving tall dense linear programs in nearly linear time.
\newblock In \emph{Proceedings of the 52nd Annual ACM SIGACT Symposium on
  Theory of Computing}, pp.\  775--788, 2020{\natexlab{b}}.

\bibitem[Brand et~al.(2024)Brand, Song, and Zhou]{bsz23}
Brand, J. v.~d., Song, Z., and Zhou, T.
\newblock Algorithm and hardness for dynamic attention maintenance in large
  language models.
\newblock In \emph{ICML}, 2024.

\bibitem[Brock et~al.(2019)Brock, Donahue, and Simonyan]{bds19}
Brock, A., Donahue, J., and Simonyan, K.
\newblock Large scale {GAN} training for high fidelity natural image synthesis.
\newblock In \emph{International Conference on Learning Representations}, 2019.
\newblock URL \url{https://openreview.net/forum?id=B1xsqj09Fm}.

\bibitem[Brown et~al.(2020)Brown, Mann, Ryder, Subbiah, Kaplan, Dhariwal,
  Neelakantan, Shyam, Sastry, Askell, et~al.]{bmr+20}
Brown, T., Mann, B., Ryder, N., Subbiah, M., Kaplan, J.~D., Dhariwal, P.,
  Neelakantan, A., Shyam, P., Sastry, G., Askell, A., et~al.
\newblock Language models are few-shot learners.
\newblock \emph{Advances in neural information processing systems},
  33:\penalty0 1877--1901, 2020.

\bibitem[Bubeck et~al.(2023)Bubeck, Chandrasekaran, Eldan, Gehrke, Horvitz,
  Kamar, Lee, Lee, Li, Lundberg, et~al.]{bce+23}
Bubeck, S., Chandrasekaran, V., Eldan, R., Gehrke, J., Horvitz, E., Kamar, E.,
  Lee, P., Lee, Y.~T., Li, Y., Lundberg, S., et~al.
\newblock Sparks of artificial general intelligence: Early experiments with
  gpt-4.
\newblock \emph{arXiv preprint arXiv:2303.12712}, 2023.

\bibitem[Burns et~al.(2023)Burns, Ye, Klein, and Steinhardt]{byks22}
Burns, C., Ye, H., Klein, D., and Steinhardt, J.
\newblock Discovering latent knowledge in language models without supervision.
\newblock \emph{ICLR}, 2023.

\bibitem[Cai et~al.(2021)Cai, Lou, McKenzie, and Yin]{clmy21}
Cai, H., Lou, Y., McKenzie, D., and Yin, W.
\newblock A zeroth-order block coordinate descent algorithm for huge-scale
  black-box optimization.
\newblock In \emph{International Conference on Machine Learning}, pp.\
  1193--1203. PMLR, 2021.

\bibitem[Cao(2024)]{c24}
Cao, Y.
\newblock Sorsa: Singular values and orthonormal regularized singular vectors
  adaptation of large language models.
\newblock \emph{arXiv preprint arXiv:2409.00055}, 2024.

\bibitem[Cao et~al.(2024)Cao, Li, and Song]{cls24}
Cao, Y., Li, X., and Song, Z.
\newblock Grams: Gradient descent with adaptive momentum scaling.
\newblock \emph{arXiv preprint arXiv:2412.17107}, 2024.

\bibitem[Cao et~al.(2025{\natexlab{a}})Cao, Chen, Li, Liang, Sha, Shi, Song,
  and Wan]{ccl+25}
Cao, Y., Chen, B., Li, X., Liang, Y., Sha, Z., Shi, Z., Song, Z., and Wan, M.
\newblock Force matching with relativistic constraints: A physics-inspired
  approach to stable and efficient generative modeling.
\newblock \emph{arXiv preprint arXiv:2502.08150}, 2025{\natexlab{a}}.

\bibitem[Cao et~al.(2025{\natexlab{b}})Cao, Guo, Huo, Liang, Shi, Song, Zhang,
  and Zhuang]{cgh+25}
Cao, Y., Guo, X., Huo, J., Liang, Y., Shi, Z., Song, Z., Zhang, J., and Zhuang,
  Z.
\newblock Text-to-image diffusion models cannot count, and prompt refinement
  cannot help.
\newblock \emph{arXiv preprint arXiv:2503.06884}, 2025{\natexlab{b}}.

\bibitem[Cao et~al.(2025{\natexlab{c}})Cao, Li, Liang, Sha, Shi, Song, and
  Zhang]{cll+25}
Cao, Y., Li, X., Liang, Y., Sha, Z., Shi, Z., Song, Z., and Zhang, J.
\newblock Dissecting submission limit in desk-rejections: A mathematical
  analysis of fairness in ai conference policies.
\newblock \emph{arXiv preprint arXiv:2502.00690}, 2025{\natexlab{c}}.

\bibitem[Cao et~al.(2025{\natexlab{d}})Cao, Song, and Yang]{csy25}
Cao, Y., Song, Z., and Yang, C.
\newblock Video latent flow matching: Optimal polynomial projections for video
  interpolation and extrapolation.
\newblock \emph{arXiv preprint arXiv:2502.00500}, 2025{\natexlab{d}}.

\bibitem[Chan et~al.(2014)Chan, Diakonikolas, Valiant, and
  Valiant]{chan2014optimal}
Chan, S.-O., Diakonikolas, I., Valiant, P., and Valiant, G.
\newblock Optimal algorithms for testing closeness of discrete distributions.
\newblock In \emph{Proceedings of the twenty-fifth annual ACM-SIAM symposium on
  Discrete algorithms}, pp.\  1193--1203. SIAM, 2014.

\bibitem[Chang et~al.(2024)Chang, Hu, Li, Yang, Jiang, and Sun]{chl+24_gat}
Chang, Y.-T., Hu, Z., Li, X., Yang, S., Jiang, J., and Sun, N.
\newblock Dihan: A novel dynamic hierarchical graph attention network for fake
  news detection.
\newblock In \emph{Proceedings of the 33rd ACM International Conference on
  Information and Knowledge Management}, pp.\  197--206, 2024.

\bibitem[ChatGPT(2022)]{cha22}
ChatGPT.
\newblock Optimizing language models for dialogue.
\newblock \emph{OpenAI Blog}, November 2022.
\newblock URL \url{https://openai.com/blog/chatgpt/}.

\bibitem[Chen et~al.(2024)Chen, Li, Liang, Long, Shi, and Song]{cll+24_rope}
Chen, B., Li, X., Liang, Y., Long, J., Shi, Z., and Song, Z.
\newblock Circuit complexity bounds for rope-based transformer architecture.
\newblock \emph{arXiv preprint arXiv:2411.07602}, 2024.

\bibitem[Chen et~al.(2025{\natexlab{a}})Chen, Shi, Song, and Zhang]{cssz25}
Chen, B., Shi, Z., Song, Z., and Zhang, J.
\newblock Provable failure of language models in learning majority boolean
  logic via gradient descent.
\newblock \emph{arXiv preprint arXiv:2504.04702}, 2025{\natexlab{a}}.

\bibitem[Chen et~al.(2023{\natexlab{a}})Chen, Kol, Paramonov, Saxena, Song, and
  Yu]{ckp+23}
Chen, L., Kol, G., Paramonov, D., Saxena, R.~R., Song, Z., and Yu, H.
\newblock Towards multi-pass streaming lower bounds for optimal approximation
  of max-cut.
\newblock In \emph{Proceedings of the 2023 Annual ACM-SIAM Symposium on
  Discrete Algorithms (SODA)}, pp.\  878--924. SIAM, 2023{\natexlab{a}}.

\bibitem[Chen et~al.(2023{\natexlab{b}})Chen, Wang, Jiang, Shi, and Xu]{cwj+23}
Chen, Y., Wang, R., Jiang, H., Shi, S., and Xu, R.
\newblock Exploring the use of large language models for reference-free text
  quality evaluation: An empirical study.
\newblock In \emph{Findings of the Association for Computational Linguistics:
  IJCNLP-AACL 2023 (Findings)}, pp.\  361--374, 2023{\natexlab{b}}.

\bibitem[Chen et~al.(2025{\natexlab{b}})Chen, Li, Liang, Shi, and
  Song]{cll+25_mamba}
Chen, Y., Li, X., Liang, Y., Shi, Z., and Song, Z.
\newblock The computational limits of state-space models and mamba via the lens
  of circuit complexity.
\newblock In \emph{Conference on Parsimony and Learning}. PMLR,
  2025{\natexlab{b}}.

\bibitem[Chervenak et~al.(2023)Chervenak, Lieman, Blanco-Breindel, and
  Jindal]{clbj23}
Chervenak, J., Lieman, H., Blanco-Breindel, M., and Jindal, S.
\newblock The promise and peril of using a large language model to obtain
  clinical information: Chatgpt performs strongly as a fertility counseling
  tool with limitations.
\newblock \emph{Fertility and Sterility}, 2023.

\bibitem[Chia et~al.(2023)Chia, Hong, Bing, and Poria]{chbp23}
Chia, Y.~K., Hong, P., Bing, L., and Poria, S.
\newblock Instructeval: Towards holistic evaluation of instruction-tuned large
  language models.
\newblock \emph{arXiv preprint arXiv:2306.04757}, 2023.

\bibitem[Choi et~al.(2023)Choi, Pei, Kumar, Shu, and Jurgens]{cpk+23}
Choi, M., Pei, J., Kumar, S., Shu, C., and Jurgens, D.
\newblock Do llms understand social knowledge? evaluating the sociability of
  large language models with socket benchmark.
\newblock In \emph{Proceedings of the 2023 Conference on Empirical Methods in
  Natural Language Processing}, pp.\  11370--11403, 2023.

\bibitem[Clarkson \& Woodruff(2013)Clarkson and Woodruff]{cw13}
Clarkson, K.~L. and Woodruff, D.~P.
\newblock Low-rank approximation and regression in input sparsity time.
\newblock In \emph{STOC}, 2013.

\bibitem[Cohen \& Peng(2015)Cohen and Peng]{cp15}
Cohen, M.~B. and Peng, R.
\newblock Lp row sampling by lewis weights.
\newblock In \emph{Proceedings of the forty-seventh annual ACM symposium on
  Theory of computing}, pp.\  183--192, 2015.

\bibitem[Cohen et~al.(2019{\natexlab{a}})Cohen, Cousins, Lee, and Yang]{ccly19}
Cohen, M.~B., Cousins, B., Lee, Y.~T., and Yang, X.
\newblock A near-optimal algorithm for approximating the john ellipsoid.
\newblock In \emph{Conference on Learning Theory}, pp.\  849--873. PMLR,
  2019{\natexlab{a}}.

\bibitem[Cohen et~al.(2019{\natexlab{b}})Cohen, Lee, and Song]{cls19}
Cohen, M.~B., Lee, Y.~T., and Song, Z.
\newblock Solving linear programs in the current matrix multiplication time.
\newblock In \emph{STOC}, 2019{\natexlab{b}}.

\bibitem[Corbett-Davies et~al.(2023)Corbett-Davies, Gaebler, Nilforoshan,
  Shroff, and Goel]{cgn+23}
Corbett-Davies, S., Gaebler, J.~D., Nilforoshan, H., Shroff, R., and Goel, S.
\newblock The measure and mismeasure of fairness.
\newblock \emph{Journal of Machine Learning Research}, 24\penalty0
  (312):\penalty0 1--117, 2023.

\bibitem[Dai et~al.(2022)Dai, Dong, Hao, Sui, Chang, and Wei]{ddh+21}
Dai, D., Dong, L., Hao, Y., Sui, Z., Chang, B., and Wei, F.
\newblock Knowledge neurons in pretrained transformers.
\newblock In \emph{Proceedings of the 60th Annual Meeting of the Association
  for Computational Linguistics (Volume 1: Long Papers)}, pp.\  8493--8502,
  2022.

\bibitem[Daitch \& Spielman(2008)Daitch and Spielman]{ds08}
Daitch, S.~I. and Spielman, D.~A.
\newblock Faster approximate lossy generalized flow via interior point
  algorithms.
\newblock In \emph{Proceedings of the fortieth annual ACM symposium on Theory
  of computing}, pp.\  451--460, 2008.

\bibitem[Das et~al.(2025)Das, Amini, and Wu]{daw25}
Das, B.~C., Amini, M.~H., and Wu, Y.
\newblock Security and privacy challenges of large language models: A survey.
\newblock \emph{ACM Computing Surveys}, 57\penalty0 (6):\penalty0 1--39, 2025.

\bibitem[Deng et~al.(2023{\natexlab{a}})Deng, Li, and Song]{dls23}
Deng, Y., Li, Z., and Song, Z.
\newblock Attention scheme inspired softmax regression.
\newblock \emph{arXiv preprint arXiv:2304.10411}, 2023{\natexlab{a}}.

\bibitem[Deng et~al.(2023{\natexlab{b}})Deng, Mahadevan, and Song]{dms23}
Deng, Y., Mahadevan, S., and Song, Z.
\newblock Randomized and deterministic attention sparsification algorithms for
  over-parameterized feature dimension.
\newblock \emph{arxiv preprint: arxiv 2304.03426}, 2023{\natexlab{b}}.

\bibitem[Deng et~al.(2024)Deng, Li, Mahadevan, and Song]{dlms23}
Deng, Y., Li, Z., Mahadevan, S., and Song, Z.
\newblock Zero-th order algorithm for softmax attention optimization.
\newblock In \emph{2024 IEEE International Conference on Big Data (BigData)},
  pp.\  24--33. IEEE, 2024.

\bibitem[Deng et~al.(2025)Deng, Li, Song, and Weinstein]{dlsw25}
Deng, Y., Li, X., Song, Z., and Weinstein, O.
\newblock Discrepancy minimization in input-sparsity time.
\newblock In \emph{Proceedings of the 42nd International Conference on Machine
  Learning (ICML)}, 2025.

\bibitem[Deroy et~al.(2023)Deroy, Ghosh, and Ghosh]{dgg23}
Deroy, A., Ghosh, K., and Ghosh, S.
\newblock How ready are pre-trained abstractive models and llms for legal case
  judgement summarization?
\newblock \emph{arXiv preprint arXiv:2306.01248}, 2023.

\bibitem[Devlin et~al.(2019)Devlin, Chang, Lee, and Toutanova]{dclt18}
Devlin, J., Chang, M.-W., Lee, K., and Toutanova, K.
\newblock Bert: Pre-training of deep bidirectional transformers for language
  understanding.
\newblock In \emph{Proceedings of the 2019 conference of the North American
  chapter of the association for computational linguistics: human language
  technologies, volume 1 (long and short papers)}, pp.\  4171--4186, 2019.

\bibitem[Dosovitskiy et~al.(2020)Dosovitskiy, Beyer, Kolesnikov, Weissenborn,
  Zhai, Unterthiner, Dehghani, Minderer, Heigold, Gelly, et~al.]{dbk+20}
Dosovitskiy, A., Beyer, L., Kolesnikov, A., Weissenborn, D., Zhai, X.,
  Unterthiner, T., Dehghani, M., Minderer, M., Heigold, G., Gelly, S., et~al.
\newblock An image is worth 16x16 words: Transformers for image recognition at
  scale.
\newblock \emph{arXiv preprint arXiv:2010.11929}, 2020.

\bibitem[Drineas et~al.(2012)Drineas, Magdon-Ismail, Mahoney, and
  Woodruff]{DMIMW12}
Drineas, P., Magdon-Ismail, M., Mahoney, M.~W., and Woodruff, D.~P.
\newblock Fast approximation of matrix coherence and statistical leverage.
\newblock \emph{The Journal of Machine Learning Research}, 13\penalty0
  (1):\penalty0 3475--3506, 2012.

\bibitem[Ferrara(2023)]{f23}
Ferrara, E.
\newblock Should chatgpt be biased? challenges and risks of bias in large
  language models.
\newblock \emph{arXiv preprint arXiv:2304.03738}, 2023.

\bibitem[Fu et~al.(2021)Fu, Liu, Zhang, Wang, Yang, Xu, Qi, Fu, and
  Zhou]{flz+21}
Fu, Z., Liu, F., Zhang, J., Wang, H., Yang, C., Xu, Q., Qi, J., Fu, X., and
  Zhou, A.
\newblock Sagn: semantic adaptive graph network for skeleton-based human action
  recognition.
\newblock In \emph{Proceedings of the 2021 International Conference on
  Multimedia Retrieval}, pp.\  110--117, 2021.

\bibitem[Gajjar et~al.(2024)Gajjar, Tai, Xingyu, Hegde, Musco, and Li]{gtx+24}
Gajjar, A., Tai, W.~M., Xingyu, X., Hegde, C., Musco, C., and Li, Y.
\newblock Agnostic active learning of single index models with linear sample
  complexity.
\newblock In \emph{The Thirty Seventh Annual Conference on Learning Theory},
  pp.\  1715--1754. PMLR, 2024.

\bibitem[Gao et~al.(2021)Gao, Liu, Zhang, Han, Liu, Niu, and Sugiyama]{glz+21}
Gao, R., Liu, F., Zhang, J., Han, B., Liu, T., Niu, G., and Sugiyama, M.
\newblock Maximum mean discrepancy test is aware of adversarial attacks.
\newblock In \emph{International Conference on Machine Learning}, pp.\
  3564--3575. PMLR, 2021.

\bibitem[Gao et~al.(2023)Gao, Mahadevan, and Song]{gms23}
Gao, Y., Mahadevan, S., and Song, Z.
\newblock An over-parameterized exponential regression.
\newblock \emph{arXiv preprint arXiv:2303.16504}, 2023.

\bibitem[Gily{\'e}n et~al.(2022)Gily{\'e}n, Song, and Tang]{gst22}
Gily{\'e}n, A., Song, Z., and Tang, E.
\newblock An improved quantum-inspired algorithm for linear regression.
\newblock \emph{Quantum}, 6:\penalty0 754, 2022.

\bibitem[Go et~al.(2022)Go, Sari, Jiang, Yang, and Jha]{gsj+22}
Go, J.~H., Sari, A., Jiang, J., Yang, S., and Jha, S.
\newblock Fake news quick detection on dynamic heterogeneous information
  networks.
\newblock \emph{arXiv preprint arXiv:2205.07039}, 2022.

\bibitem[Goldreich \& Ron(2011)Goldreich and Ron]{goldreich2011testing}
Goldreich, O. and Ron, D.
\newblock On testing expansion in bounded-degree graphs.
\newblock \emph{Studies in Complexity and Cryptography. Miscellanea on the
  Interplay between Randomness and Computation: In Collaboration with Lidor
  Avigad, Mihir Bellare, Zvika Brakerski, Shafi Goldwasser, Shai Halevi, Tali
  Kaufman, Leonid Levin, Noam Nisan, Dana Ron, Madhu Sudan, Luca Trevisan,
  Salil Vadhan, Avi Wigderson, David Zuckerman}, pp.\  68--75, 2011.

\bibitem[Gu et~al.(2024)Gu, Song, Yin, and Zhang]{gsyz23}
Gu, Y., Song, Z., Yin, J., and Zhang, L.
\newblock Low rank matrix completion via robust alternating minimization in
  nearly linear time.
\newblock In \emph{ICLR}, 2024.

\bibitem[Gu et~al.(2025)Gu, Song, and Zhang]{gsz25}
Gu, Y., Song, Z., and Zhang, L.
\newblock Faster algorithms for structured linear and kernel support vector
  machines.
\newblock In \emph{ICLR}, 2025.

\bibitem[Guo et~al.(2023)Guo, Guo, Liang, Guo, Chawla, Wiest, Zhang,
  et~al.]{ggl+23}
Guo, T., Guo, K., Liang, Z., Guo, Z., Chawla, N.~V., Wiest, O., Zhang, X.,
  et~al.
\newblock What indeed can gpt models do in chemistry? a comprehensive benchmark
  on eight tasks.
\newblock \emph{arXiv preprint arXiv:2305.18365}, 2023.

\bibitem[Guo et~al.(2025{\natexlab{a}})Guo, Huang, Huo, Liang, Shi, Song, and
  Zhang]{ghh+25}
Guo, X., Huang, Z., Huo, J., Liang, Y., Shi, Z., Song, Z., and Zhang, J.
\newblock Can you count to nine? a human evaluation benchmark for counting
  limits in modern text-to-video models.
\newblock \emph{arXiv preprint arXiv:2504.04051}, 2025{\natexlab{a}}.

\bibitem[Guo et~al.(2025{\natexlab{b}})Guo, Huo, Shi, Song, Zhang, and
  Zhao]{ghs+25_physical}
Guo, X., Huo, J., Shi, Z., Song, Z., Zhang, J., and Zhao, J.
\newblock T2vphysbench: A first-principles benchmark for physical consistency
  in text-to-video generation.
\newblock \emph{arXiv preprint arXiv:2505.00337}, 2025{\natexlab{b}}.

\bibitem[Hase et~al.(2023)Hase, Bansal, Kim, and Ghandeharioun]{hbkg23}
Hase, P., Bansal, M., Kim, B., and Ghandeharioun, A.
\newblock Does localization inform editing? surprising differences in
  causality-based localization vs. knowledge editing in language models.
\newblock \emph{Advances in Neural Information Processing Systems},
  36:\penalty0 17643--17668, 2023.

\bibitem[Hewitt \& Manning(2019)Hewitt and Manning]{hm19}
Hewitt, J. and Manning, C.~D.
\newblock A structural probe for finding syntax in word representations.
\newblock In \emph{Proceedings of the 2019 Conference of the North American
  Chapter of the Association for Computational Linguistics: Human Language
  Technologies, Volume 1 (Long and Short Papers)}, pp.\  4129--4138, 2019.

\bibitem[Hu et~al.(2022)Hu, Shen, Wallis, Allen-Zhu, Li, Wang, Wang, Chen,
  et~al.]{hsw+22}
Hu, E.~J., Shen, Y., Wallis, P., Allen-Zhu, Z., Li, Y., Wang, S., Wang, L.,
  Chen, W., et~al.
\newblock Lora: Low-rank adaptation of large language models.
\newblock \emph{ICLR}, 1\penalty0 (2):\penalty0 3, 2022.

\bibitem[Hu et~al.(2024{\natexlab{a}})Hu, Chen, Wu, Ruan, and Liu]{hcw+24}
Hu, J. Y.-C., Chen, B.-Y., Wu, D., Ruan, F., and Liu, H.
\newblock Nonparametric modern hopfield models.
\newblock \emph{arXiv preprint arXiv:2404.03900}, 2024{\natexlab{a}}.

\bibitem[Hu et~al.(2024{\natexlab{b}})Hu, Lin, Song, and Liu]{hlsl24}
Hu, J. Y.-C., Lin, T., Song, Z., and Liu, H.
\newblock On computational limits of modern hopfield models: A fine-grained
  complexity analysis.
\newblock In \emph{Forty-first International Conference on Machine Learning},
  2024{\natexlab{b}}.

\bibitem[Hu et~al.(2024{\natexlab{c}})Hu, Su, Kuo, Song, and Liu]{hsk+24}
Hu, J. Y.-C., Su, M., Kuo, E.-J., Song, Z., and Liu, H.
\newblock Computational limits of low-rank adaptation (lora) for
  transformer-based models.
\newblock \emph{arXiv preprint arXiv:2406.03136}, 2024{\natexlab{c}}.

\bibitem[Hu et~al.(2024{\natexlab{d}})Hu, Wu, and Liu]{hwl24}
Hu, J. Y.-C., Wu, D., and Liu, H.
\newblock Provably optimal memory capacity for modern hopfield models:
  Transformer-compatible dense associative memories as spherical codes.
\newblock In \emph{Thirty-eighth Conference on Neural Information Processing
  Systems (NeurIPS)}, 2024{\natexlab{d}}.

\bibitem[Hu et~al.(2025)Hu, Zhang, Su, Song, and Liu]{hzs+25}
Hu, J. Y.-C., Zhang, X., Su, M., Song, Z., and Liu, H.
\newblock Minimalist softmax attention provably learns constrained boolean
  functions.
\newblock \emph{arXiv preprint arXiv:2505.19531}, 2025.

\bibitem[Huang et~al.(2022)Huang, Jiang, Song, Tao, and Zhang]{hjs+22}
Huang, B., Jiang, S., Song, Z., Tao, R., and Zhang, R.
\newblock Solving sdp faster: A robust ipm framework and efficient
  implementation.
\newblock In \emph{2022 IEEE 63rd Annual Symposium on Foundations of Computer
  Science (FOCS)}, pp.\  233--244. IEEE, 2022.

\bibitem[Huang et~al.(2025)Huang, Sun, Tani, Zhang, Jiang, and Jha]{hst+25}
Huang, H., Sun, N., Tani, M., Zhang, Y., Jiang, J., and Jha, S.
\newblock Can llm-generated misinformation be detected: A study on cyber threat
  intelligence.
\newblock \emph{Future Generation Computer Systems}, pp.\  107877, 2025.

\bibitem[Huang et~al.(2023)Huang, Liu, Du, and Tao]{hldt23}
Huang, X., Liu, W., Du, B., and Tao, D.
\newblock Leveraged matrix completion with noise.
\newblock \emph{IEEE Transactions on Cybernetics}, 54\penalty0 (8):\penalty0
  4443--4453, 2023.

\bibitem[Ingster(1982)]{ingster1982minimax}
Ingster, Y.~I.
\newblock On the minimax nonparametric detection of signals in white {G}aussian
  noise.
\newblock \emph{Problemy Peredachi Informatsii}, 18\penalty0 (2):\penalty0
  61--73, 1982.

\bibitem[Ingster(1987)]{ingster1987minimax}
Ingster, Y.~I.
\newblock Minimax testing of nonparametric hypotheses on a distribution density
  in the l\_p metrics.
\newblock \emph{Theory of Probability \& Its Applications}, 31\penalty0
  (2):\penalty0 333--337, 1987.

\bibitem[Ioffe \& Szegedy(2015)Ioffe and Szegedy]{ioffe2015batch}
Ioffe, S. and Szegedy, C.
\newblock Batch normalization: Accelerating deep network training by reducing
  internal covariate shift.
\newblock In \emph{International Conference on Machine Learning (ICML)}, pp.\
  448--456. PMLR, 2015.

\bibitem[Jiang et~al.(2020{\natexlab{a}})Jiang, Kathuria, Lee, Padmanabhan, and
  Song]{jkl+20}
Jiang, H., Kathuria, T., Lee, Y.~T., Padmanabhan, S., and Song, Z.
\newblock A faster interior point method for semidefinite programming.
\newblock In \emph{2020 IEEE 61st annual symposium on foundations of computer
  science (FOCS)}, pp.\  910--918. IEEE, 2020{\natexlab{a}}.

\bibitem[Jiang et~al.(2020{\natexlab{b}})Jiang, Lee, Song, and Wong]{jlsw20}
Jiang, H., Lee, Y.~T., Song, Z., and Wong, S. C.-w.
\newblock An improved cutting plane method for convex optimization,
  convex-concave games and its applications.
\newblock In \emph{STOC}, 2020{\natexlab{b}}.

\bibitem[Johnson et~al.(2023)Johnson, Goodman, Patrinely, Stone, Zimmerman,
  Donald, Chang, Berkowitz, Finn, Jahangir, et~al.]{jgp+23}
Johnson, D., Goodman, R., Patrinely, J., Stone, C., Zimmerman, E., Donald, R.,
  Chang, S., Berkowitz, S., Finn, A., Jahangir, E., et~al.
\newblock Assessing the accuracy and reliability of ai-generated medical
  responses: an evaluation of the chat-gpt model.
\newblock \emph{Research square}, pp.\  rs--3, 2023.

\bibitem[Kang \& McAuley(2018)Kang and McAuley]{km18}
Kang, W.-C. and McAuley, J.
\newblock Self-attentive sequential recommendation.
\newblock In \emph{ICDM}, 2018.

\bibitem[Kaplan et~al.(2020)Kaplan, McCandlish, Henighan, Brown, Chess, Child,
  Gray, Radford, Wu, and Amodei]{kmh+20}
Kaplan, J., McCandlish, S., Henighan, T., Brown, T.~B., Chess, B., Child, R.,
  Gray, S., Radford, A., Wu, J., and Amodei, D.
\newblock Scaling laws for neural language models.
\newblock \emph{arXiv preprint arXiv:2001.08361}, 2020.

\bibitem[Karunanayake et~al.(2023)Karunanayake, Jiang, Ahmed, and Jha]{kja+23}
Karunanayake, I., Jiang, J., Ahmed, N., and Jha, S.~K.
\newblock Exploring uncharted waters of website fingerprinting.
\newblock \emph{IEEE Transactions on Information Forensics and Security},
  19:\penalty0 1840--1854, 2023.

\bibitem[Ke et~al.(2025)Ke, Li, Liang, Shi, and Song]{kll+25_tc}
Ke, Y., Li, X., Liang, Y., Shi, Z., and Song, Z.
\newblock Circuit complexity bounds for visual autoregressive model.
\newblock \emph{arXiv preprint arXiv:2501.04299}, 2025.

\bibitem[Kim \& Suzuki(2025)Kim and Suzuki]{ks25}
Kim, J. and Suzuki, T.
\newblock Transformers provably solve parity efficiently with chain of thought.
\newblock In \emph{ICLR}, 2025.

\bibitem[Laskar et~al.(2023)Laskar, Bari, Rahman, Bhuiyan, Joty, and
  Huang]{lbr+23}
Laskar, M. T.~R., Bari, M.~S., Rahman, M., Bhuiyan, M. A.~H., Joty, S., and
  Huang, J.
\newblock A systematic study and comprehensive evaluation of chatgpt on
  benchmark datasets.
\newblock In \emph{Findings of the Association for Computational Linguistics:
  ACL 2023}, pp.\  431--469, 2023.

\bibitem[Lee et~al.(2020)Lee, Shen, Song, Wang, et~al.]{lls+20}
Lee, J.~D., Shen, R., Song, Z., Wang, M., et~al.
\newblock Generalized leverage score sampling for neural networks.
\newblock \emph{Advances in Neural Information Processing Systems},
  33:\penalty0 10775--10787, 2020.

\bibitem[Lee \& Sidford(2014)Lee and Sidford]{ls14}
Lee, Y.~T. and Sidford, A.
\newblock Path finding methods for linear programming: Solving linear programs
  in o (vrank) iterations and faster algorithms for maximum flow.
\newblock In \emph{2014 IEEE 55th Annual Symposium on Foundations of Computer
  Science}, pp.\  424--433. IEEE, 2014.

\bibitem[Lee et~al.(2015)Lee, Sidford, and Wong]{lsw15}
Lee, Y.~T., Sidford, A., and Wong, S. C.-w.
\newblock A faster cutting plane method and its implications for combinatorial
  and convex optimization.
\newblock In \emph{2015 IEEE 56th Annual Symposium on Foundations of Computer
  Science}, pp.\  1049--1065. IEEE, 2015.

\bibitem[Lee et~al.(2019)Lee, Song, and Zhang]{lsz19}
Lee, Y.~T., Song, Z., and Zhang, Q.
\newblock Solving empirical risk minimization in the current matrix
  multiplication time.
\newblock In \emph{Conference on Learning Theory}, pp.\  2140--2157. PMLR,
  2019.

\bibitem[Li \& Yuan(2019)Li and Yuan]{li2019optimality}
Li, T. and Yuan, M.
\newblock On the optimality of {G}aussian kernel based nonparametric tests
  against smooth alternatives.
\newblock \emph{arXiv preprint arXiv:1909.03302}, 2019.

\bibitem[Li et~al.(2024{\natexlab{a}})Li, Liang, Shi, Song, and Yu]{lls+24}
Li, X., Liang, Y., Shi, Z., Song, Z., and Yu, J.
\newblock Fast john ellipsoid computation with differential privacy
  optimization.
\newblock \emph{arXiv preprint arXiv:2408.06395}, 2024{\natexlab{a}}.

\bibitem[Li et~al.(2024{\natexlab{b}})Li, Song, and Yu]{lsy24}
Li, X., Song, Z., and Yu, J.
\newblock Quantum speedups for approximating the john ellipsoid.
\newblock \emph{arXiv preprint arXiv:2408.14018}, 2024{\natexlab{b}}.

\bibitem[Li et~al.(2025{\natexlab{a}})Li, Liang, Shi, Song, Wang, and
  Zhang]{lls+25}
Li, X., Liang, Y., Shi, Z., Song, Z., Wang, W., and Zhang, J.
\newblock On the computational capability of graph neural networks: A circuit
  complexity bound perspective.
\newblock \emph{arXiv preprint arXiv:2501.06444}, 2025{\natexlab{a}}.

\bibitem[Li et~al.(2025{\natexlab{b}})Li, Sun, and Jiang]{lsj25}
Li, X., Sun, N., and Jiang, J.
\newblock Llm-based approaches for real-time cyber threat detection and
  response: A survey.
\newblock \emph{manuscript}, 2025{\natexlab{b}}.

\bibitem[Li et~al.(2025{\natexlab{c}})Li, Xie, and Song]{lxz25}
Li, X., Xie, S., and Song, Z.
\newblock Deterministic sparse fourier transform for continuous signals with
  frequency gap.
\newblock In \emph{Proceedings of the 42nd International Conference on Machine
  Learning (ICML)}, 2025{\natexlab{c}}.

\bibitem[Li \& Liang(2021)Li and Liang]{ll21}
Li, X.~L. and Liang, P.
\newblock Prefix-tuning: Optimizing continuous prompts for generation.
\newblock In \emph{ACL}, 2021.

\bibitem[Li et~al.(2023)Li, You, Bhojanapalli, Li, Rawat, Reddi, Ye, Chern, Yu,
  Guo, et~al.]{lyb+22}
Li, Z., You, C., Bhojanapalli, S., Li, D., Rawat, A.~S., Reddi, S.~J., Ye, K.,
  Chern, F., Yu, F., Guo, R., et~al.
\newblock The lazy neuron phenomenon: On emergence of activation sparsity in
  transformers.
\newblock \emph{ICLR}, 2023.

\bibitem[Li et~al.(2024{\natexlab{c}})Li, Liu, Zhou, and Ma]{llzm24}
Li, Z., Liu, H., Zhou, D., and Ma, T.
\newblock Chain of thought empowers transformers to solve inherently serial
  problems.
\newblock In \emph{ICLR}, 2024{\natexlab{c}}.

\bibitem[Liang et~al.(2023)Liang, Bommasani, Lee, Tsipras, Soylu, Yasunaga,
  Zhang, Narayanan, Wu, Kumar, et~al.]{lbl+22}
Liang, P., Bommasani, R., Lee, T., Tsipras, D., Soylu, D., Yasunaga, M., Zhang,
  Y., Narayanan, D., Wu, Y., Kumar, A., et~al.
\newblock Holistic evaluation of language models.
\newblock \emph{Transactions on Machine Learning Research (TMLR)}, 2023.

\bibitem[Lipman et~al.(2022)Lipman, Chen, Ben-Hamu, Nickel, and Le]{lcb+22}
Lipman, Y., Chen, R.~T., Ben-Hamu, H., Nickel, M., and Le, M.
\newblock Flow matching for generative modeling.
\newblock \emph{arXiv preprint arXiv:2210.02747}, 2022.

\bibitem[Liu et~al.(2024{\natexlab{a}})Liu, Feng, Xue, Wang, Wu, Lu, Zhao,
  Deng, Zhang, Ruan, et~al.]{deepseekv3}
Liu, A., Feng, B., Xue, B., Wang, B., Wu, B., Lu, C., Zhao, C., Deng, C.,
  Zhang, C., Ruan, C., et~al.
\newblock Deepseek-v3 technical report.
\newblock \emph{arXiv preprint arXiv:2412.19437}, 2024{\natexlab{a}}.

\bibitem[Liu et~al.(2024{\natexlab{b}})Liu, Zhang, Wang, Fan, and Li]{lzw+24}
Liu, C., Zhang, J., Wang, S., Fan, W., and Li, Q.
\newblock Score-based generative diffusion models for social recommendations.
\newblock \emph{arXiv preprint arXiv:2412.15579}, 2024{\natexlab{b}}.

\bibitem[Liu et~al.(2020{\natexlab{a}})Liu, Xu, Lu, Zhang, Gretton, and
  Sutherland]{lxl+20}
Liu, F., Xu, W., Lu, J., Zhang, G., Gretton, A., and Sutherland, D.~J.
\newblock Learning deep kernels for non-parametric two-sample tests.
\newblock In \emph{International conference on machine learning}, pp.\
  6316--6326. PMLR, 2020{\natexlab{a}}.

\bibitem[Liu et~al.(2021{\natexlab{a}})Liu, Xu, Lu, and Sutherland]{lxls21}
Liu, F., Xu, W., Lu, J., and Sutherland, D.~J.
\newblock Meta two-sample testing: Learning kernels for testing with limited
  data.
\newblock \emph{Advances in Neural Information Processing Systems},
  34:\penalty0 5848--5860, 2021{\natexlab{a}}.

\bibitem[Liu et~al.(2022{\natexlab{a}})Liu, Wang, Zhang, Fu, Zhou, Qi, and
  Li]{lwz+22}
Liu, F., Wang, H., Zhang, J., Fu, Z., Zhou, A., Qi, J., and Li, Z.
\newblock Evogan: An evolutionary computation assisted gan.
\newblock \emph{Neurocomputing}, 469:\penalty0 81--90, 2022{\natexlab{a}}.

\bibitem[Liu et~al.(2024{\natexlab{c}})Liu, Li, Hall, Liang, and Ma]{llh+23}
Liu, H., Li, Z., Hall, D., Liang, P., and Ma, T.
\newblock Sophia: A scalable stochastic second-order optimizer for language
  model pre-training.
\newblock \emph{ICLR}, 2024{\natexlab{c}}.

\bibitem[Liu et~al.(2023{\natexlab{a}})Liu, Xia, Wang, and Zhang]{lxwz23}
Liu, J., Xia, C.~S., Wang, Y., and Zhang, L.
\newblock Is your code generated by chatgpt really correct? rigorous evaluation
  of large language models for code generation.
\newblock \emph{NeurIPS}, 2023{\natexlab{a}}.

\bibitem[Liu et~al.(2023{\natexlab{b}})Liu, Cai, Zhang, Zhao, Gao, Wang, Lv,
  Fan, Wang, He, Liu, and Li]{llz+23}
Liu, L., Cai, L., Zhang, C., Zhao, X., Gao, J., Wang, W., Lv, Y., Fan, W.,
  Wang, Y., He, M., Liu, Z., and Li, Q.
\newblock Linrec: Linear attention mechanism for long-term sequential
  recommender systems.
\newblock In \emph{SIGIR}, 2023{\natexlab{b}}.

\bibitem[Liu et~al.(2020{\natexlab{b}})Liu, Song, and Zhang]{lsz20}
Liu, S.~C., Song, Z., and Zhang, H.
\newblock Breaking the n-pass barrier: A streaming algorithm for maximum weight
  bipartite matching.
\newblock \emph{arXiv preprint arXiv:2009.06106}, 2020{\natexlab{b}}.

\bibitem[Liu et~al.(2023{\natexlab{c}})Liu, Song, Zhang, Zhang, and
  Zhou]{lsz+22}
Liu, S.~C., Song, Z., Zhang, H., Zhang, L., and Zhou, T.
\newblock Space-efficient interior point method, with applications to linear
  programming and maximum weight bipartite matching.
\newblock In \emph{50th International Colloquium on Automata, Languages, and
  Programming (ICALP 2023)}, pp.\  88--1. Schloss Dagstuhl--Leibniz-Zentrum
  f{\"u}r Informatik, 2023{\natexlab{c}}.

\bibitem[Liu et~al.(2023{\natexlab{d}})Liu, Song, Zhang, Zhang, and
  Zhou]{lsz+23}
Liu, S.~C., Song, Z., Zhang, H., Zhang, L., and Zhou, T.
\newblock Space-efficient interior point method, with applications to linear
  programming and maximum weight bipartite matching.
\newblock In \emph{ICALP}, 2023{\natexlab{d}}.

\bibitem[Liu et~al.(2022{\natexlab{b}})Liu, Gong, and Liu]{lgl22}
Liu, X., Gong, C., and Liu, Q.
\newblock Flow straight and fast: Learning to generate and transfer data with
  rectified flow.
\newblock \emph{arXiv preprint arXiv:2209.03003}, 2022{\natexlab{b}}.

\bibitem[Liu \& Sidford(2020)Liu and Sidford]{ls20}
Liu, Y.~P. and Sidford, A.
\newblock Faster energy maximization for faster maximum flow.
\newblock In \emph{Proceedings of the 52nd Annual ACM SIGACT Symposium on
  Theory of Computing}, pp.\  803--814, 2020.

\bibitem[Liu et~al.(2021{\natexlab{b}})Liu, Lin, Cao, Hu, Wei, Zhang, Lin, and
  Guo]{llc+21}
Liu, Z., Lin, Y., Cao, Y., Hu, H., Wei, Y., Zhang, Z., Lin, S., and Guo, B.
\newblock Swin transformer: Hierarchical vision transformer using shifted
  windows.
\newblock In \emph{ICCV}, 2021{\natexlab{b}}.

\bibitem[Madry(2013)]{m13}
Madry, A.
\newblock Navigating central path with electrical flows: From flows to
  matchings, and back.
\newblock In \emph{2013 IEEE 54th Annual Symposium on Foundations of Computer
  Science}, pp.\  253--262. IEEE, 2013.

\bibitem[Madry(2016)]{m16}
Madry, A.
\newblock Computing maximum flow with augmenting electrical flows.
\newblock In \emph{2016 IEEE 57th Annual Symposium on Foundations of Computer
  Science (FOCS)}, pp.\  593--602. IEEE, 2016.

\bibitem[Malladi et~al.(2023)Malladi, Gao, Nichani, Damian, Lee, Chen, and
  Arora]{mgn+23}
Malladi, S., Gao, T., Nichani, E., Damian, A., Lee, J.~D., Chen, D., and Arora,
  S.
\newblock Fine-tuning language models with just forward passes.
\newblock \emph{NeurIPS}, 2023.

\bibitem[Meng et~al.(2022)Meng, Bau, Andonian, and Belinkov]{mbab22}
Meng, K., Bau, D., Andonian, A., and Belinkov, Y.
\newblock Locating and editing factual associations in gpt.
\newblock \emph{Advances in Neural Information Processing Systems},
  35:\penalty0 17359--17372, 2022.

\bibitem[Nay et~al.(2023)Nay, Karamardian, Lawsky, Tao, Bhat, Jain, Lee, Choi,
  and Kasai]{nkl+23}
Nay, J.~J., Karamardian, D., Lawsky, S.~B., Tao, W., Bhat, M., Jain, R., Lee,
  A.~T., Choi, J.~H., and Kasai, J.
\newblock Large language models as tax attorneys: A case study in legal
  capabilities emergence.
\newblock \emph{arXiv preprint arXiv:2306.07075}, 2023.

\bibitem[Neyman \& Pearson(1933)Neyman and Pearson]{neyman1933ix}
Neyman, J. and Pearson, E.~S.
\newblock Ix. on the problem of the most efficient tests of statistical
  hypotheses.
\newblock \emph{Philosophical Transactions of the Royal Society of London.
  Series A, Containing Papers of a Mathematical or Physical Character},
  231\penalty0 (694-706):\penalty0 289--337, 1933.

\bibitem[OpenAI(2023)]{o23}
OpenAI.
\newblock Gpt-4 technical report.
\newblock \emph{arXiv preprint arXiv:2303.08774}, 2023.

\bibitem[Pallagani et~al.(2023)Pallagani, Muppasani, Murugesan, Rossi,
  Srivastava, Horesh, Fabiano, and Loreggia]{pmm+23}
Pallagani, V., Muppasani, B., Murugesan, K., Rossi, F., Srivastava, B., Horesh,
  L., Fabiano, F., and Loreggia, A.
\newblock Understanding the capabilities of large language models for automated
  planning.
\newblock \emph{arXiv preprint arXiv:2305.16151}, 2023.

\bibitem[Peebles \& Xie(2023)Peebles and Xie]{px23}
Peebles, W. and Xie, S.
\newblock Scalable diffusion models with transformers.
\newblock In \emph{Proceedings of the IEEE/CVF international conference on
  computer vision}, pp.\  4195--4205, 2023.

\bibitem[Pensia et~al.(2023)Pensia, Asadi, Jog, and Loh]{pajl23}
Pensia, A., Asadi, A.~R., Jog, V., and Loh, P.-L.
\newblock Simple binary hypothesis testing under local differential privacy and
  communication constraints.
\newblock In \emph{COLT}, 2023.

\bibitem[Pensia et~al.(2024)Pensia, Jog, and Loh]{pjl24}
Pensia, A., Jog, V., and Loh, P.-L.
\newblock The sample complexity of simple binary hypothesis testing.
\newblock In \emph{COLT}, 2024.

\bibitem[Polyanskiy \& Wu(2023+)Polyanskiy and Wu]{polyanskiy2023information}
Polyanskiy, Y. and Wu, Y.
\newblock \emph{Information Theory: From Coding to Learning}.
\newblock Cambridge University Press, 2023+.

\bibitem[Price et~al.(2017)Price, Song, and Woodruff]{psw17}
Price, E., Song, Z., and Woodruff, D.~P.
\newblock Fast regression with an {$\ell_\infty$} guarantee.
\newblock In \emph{International Colloquium on Automata, Languages, and
  Programming}. Schloss Dagstuhl-Leibniz-Zentrum fur Informatik GmbH, Dagstuhl
  Publishing, 2017.

\bibitem[Pu \& Demberg(2023)Pu and Demberg]{pd23}
Pu, D. and Demberg, V.
\newblock Chatgpt vs human-authored text: Insights into controllable text
  summarization and sentence style transfer.
\newblock \emph{ACL (Student Abstract)}, 2023.

\bibitem[Qin et~al.(2023{\natexlab{a}})Qin, Zhang, Zhang, Chen, Yasunaga, and
  Yang]{qzz+23}
Qin, C., Zhang, A., Zhang, Z., Chen, J., Yasunaga, M., and Yang, D.
\newblock Is chatgpt a general-purpose natural language processing task solver?
\newblock \emph{EMNLP}, 2023{\natexlab{a}}.

\bibitem[Qin et~al.(2023{\natexlab{b}})Qin, Song, Zhang, and Zhuo]{qszz23}
Qin, L., Song, Z., Zhang, L., and Zhuo, D.
\newblock An online and unified algorithm for projection matrix vector
  multiplication with application to empirical risk minimization.
\newblock In \emph{AISTATS}, 2023{\natexlab{b}}.

\bibitem[Radford et~al.(2018)Radford, Narasimhan, Salimans, Sutskever,
  et~al.]{rns+18}
Radford, A., Narasimhan, K., Salimans, T., Sutskever, I., et~al.
\newblock Improving language understanding by generative pre-training.
\newblock \emph{OpenAI Blog}, 2018.

\bibitem[Radford et~al.(2019)Radford, Wu, Child, Luan, Amodei, Sutskever,
  et~al.]{rwc+19}
Radford, A., Wu, J., Child, R., Luan, D., Amodei, D., Sutskever, I., et~al.
\newblock Language models are unsupervised multitask learners.
\newblock \emph{OpenAI blog}, 1\penalty0 (8):\penalty0 9, 2019.

\bibitem[Rafailov et~al.(2023)Rafailov, Sharma, Mitchell, Ermon, Manning, and
  Finn]{rsm+23}
Rafailov, R., Sharma, A., Mitchell, E., Ermon, S., Manning, C.~D., and Finn, C.
\newblock Direct preference optimization: Your language model is secretly a
  reward model.
\newblock \emph{NeurIPS}, 2023.

\bibitem[Rathee(2020)]{r20}
Rathee, K.
\newblock Meet google meena, 2020.

\bibitem[Reif et~al.(2019)Reif, Yuan, Wattenberg, Viegas, Coenen, Pearce, and
  Kim]{ryw+19}
Reif, E., Yuan, A., Wattenberg, M., Viegas, F.~B., Coenen, A., Pearce, A., and
  Kim, B.
\newblock Visualizing and measuring the geometry of bert.
\newblock \emph{Advances in Neural Information Processing Systems}, 32, 2019.

\bibitem[Salemi \& Zamani(2024)Salemi and Zamani]{sz24}
Salemi, A. and Zamani, H.
\newblock Evaluating retrieval quality in retrieval-augmented generation.
\newblock \emph{SIGIR}, 2024.

\bibitem[Schild(2018)]{s18}
Schild, A.
\newblock An almost-linear time algorithm for uniform random spanning tree
  generation.
\newblock In \emph{Proceedings of the 50th Annual ACM SIGACT Symposium on
  Theory of Computing}, pp.\  214--227, 2018.

\bibitem[Shen et~al.(2025)Shen, Song, Zhou, Chen, Li, Gong, Zhang, Tan, Kuen,
  Ding, et~al.]{ssz+25}
Shen, X., Song, Z., Zhou, Y., Chen, B., Li, Y., Gong, Y., Zhang, K., Tan, H.,
  Kuen, J., Ding, H., et~al.
\newblock Lazydit: Lazy learning for the acceleration of diffusion
  transformers.
\newblock In \emph{Proceedings of the AAAI Conference on Artificial
  Intelligence}, volume 39:19, pp.\  20409--20417, 2025.

\bibitem[Shimizu et~al.(2024)Shimizu, Cheng, Musco, and Weare]{scmw24}
Shimizu, A., Cheng, X., Musco, C., and Weare, J.
\newblock Improved active learning via dependent leverage score sampling.
\newblock In \emph{The Twelfth International Conference on Learning
  Representations}, 2024.

\bibitem[Shin et~al.(2023)Shin, Shomorony, and Zhao]{ssz23_gnn}
Shin, S., Shomorony, I., and Zhao, H.
\newblock Efficient learning of linear graph neural networks via node
  subsampling.
\newblock \emph{Advances in Neural Information Processing Systems},
  36:\penalty0 55479--55501, 2023.

\bibitem[Siriwardhana et~al.(2023)Siriwardhana, Weerasekera, Wen, Kaluarachchi,
  Rana, and Nanayakkara]{sww+23}
Siriwardhana, S., Weerasekera, R., Wen, E., Kaluarachchi, T., Rana, R., and
  Nanayakkara, S.
\newblock Improving the domain adaptation of retrieval augmented generation
  (rag) models for open domain question answering.
\newblock \emph{Transactions of the Association for Computational Linguistics},
  11:\penalty0 1--17, 2023.

\bibitem[Song et~al.(2025)Song, Yuan, Zhang, Fang, Yu, and Liu]{syz+25}
Song, Y., Yuan, Z., Zhang, S., Fang, Z., Yu, J., and Liu, F.
\newblock Deep kernel relative test for machine-generated text detection.
\newblock In \emph{The Thirteenth International Conference on Learning
  Representations}, 2025.

\bibitem[Song(2019)]{s19}
Song, Z.
\newblock \emph{Matrix Theory: Optimization, Concentration and Algorithms}.
\newblock PhD thesis, The University of Texas at Austin, 2019.

\bibitem[Song \& Yu(2021)Song and Yu]{sy21}
Song, Z. and Yu, Z.
\newblock Oblivious sketching-based central path method for solving linear
  programming problems.
\newblock In \emph{38th International Conference on Machine Learning (ICML)},
  2021.

\bibitem[Song et~al.(2017)Song, Woodruff, and Zhong]{swz17}
Song, Z., Woodruff, D.~P., and Zhong, P.
\newblock Low rank approximation with entrywise l1-norm error.
\newblock In \emph{Proceedings of the 49th Annual ACM SIGACT Symposium on
  Theory of Computing}, pp.\  688--701, 2017.

\bibitem[Song et~al.(2019{\natexlab{a}})Song, Wang, Yang, Zhang, and
  Zhong]{swy+19}
Song, Z., Wang, R., Yang, L., Zhang, H., and Zhong, P.
\newblock Efficient symmetric norm regression via linear sketching.
\newblock \emph{Advances in Neural Information Processing Systems}, 32,
  2019{\natexlab{a}}.

\bibitem[Song et~al.(2019{\natexlab{b}})Song, Woodruff, and Zhong]{swz19}
Song, Z., Woodruff, D.~P., and Zhong, P.
\newblock Relative error tensor low rank approximation.
\newblock In \emph{Proceedings of the Thirtieth Annual ACM-SIAM Symposium on
  Discrete Algorithms}, pp.\  2772--2789. SIAM, 2019{\natexlab{b}}.

\bibitem[Song et~al.(2022)Song, Yang, Yang, and Zhou]{syyz22}
Song, Z., Yang, X., Yang, Y., and Zhou, T.
\newblock Faster algorithm for structured john ellipsoid computation.
\newblock \emph{arXiv preprint arXiv:2211.14407}, 2022.

\bibitem[Song et~al.(2023)Song, Ye, Yin, and Zhang]{syyz23_linf}
Song, Z., Ye, M., Yin, J., and Zhang, L.
\newblock A nearly-optimal bound for fast regression with $\ell_\infty$
  guarantee.
\newblock In \emph{International Conference on Machine Learning (ICML)}, pp.\
  32463--32482. PMLR, 2023.

\bibitem[Song et~al.(2024{\natexlab{a}})Song, Vakilian, Woodruff, and
  Zhou]{svwz24}
Song, Z., Vakilian, A., Woodruff, D., and Zhou, S.
\newblock On socially fair low-rank approximation and column subset selection.
\newblock \emph{Advances in Neural Information Processing Systems},
  37:\penalty0 88874--88905, 2024{\natexlab{a}}.

\bibitem[Song et~al.(2024{\natexlab{b}})Song, Yin, and Zhang]{syz23}
Song, Z., Yin, J., and Zhang, L.
\newblock Solving attention kernel regression problem via pre-conditioner.
\newblock \emph{AISTATS}, 2024{\natexlab{b}}.

\bibitem[Spataro(2023)]{s23}
Spataro, J.
\newblock Introducing microsoft 365 copilot – your copilot for work, 2023.

\bibitem[Spielman \& Srivastava(2008{\natexlab{a}})Spielman and
  Srivastava]{s11}
Spielman, D.~A. and Srivastava, N.
\newblock Graph sparsification by effective resistances.
\newblock In \emph{Proceedings of the fortieth annual ACM symposium on Theory
  of computing}, pp.\  563--568, 2008{\natexlab{a}}.

\bibitem[Spielman \& Srivastava(2008{\natexlab{b}})Spielman and
  Srivastava]{ss11}
Spielman, D.~A. and Srivastava, N.
\newblock Graph sparsification by effective resistances.
\newblock In \emph{Proceedings of the fortieth annual ACM symposium on Theory
  of computing}, pp.\  563--568, 2008{\natexlab{b}}.

\bibitem[Sridhara et~al.(2023)Sridhara, Mazumdar, et~al.]{sm+23}
Sridhara, G., Mazumdar, S., et~al.
\newblock Chatgpt: A study on its utility for ubiquitous software engineering
  tasks.
\newblock \emph{arXiv preprint arXiv:2305.16837}, 2023.

\bibitem[Sun et~al.(2023)Sun, Ding, Jiang, Xu, Mo, Tai, and Zhang]{sdj+23}
Sun, N., Ding, M., Jiang, J., Xu, W., Mo, X., Tai, Y., and Zhang, J.
\newblock Cyber threat intelligence mining for proactive cybersecurity defense:
  A survey and new perspectives.
\newblock \emph{IEEE Communications Surveys \& Tutorials}, 25\penalty0
  (3):\penalty0 1748--1774, 2023.

\bibitem[Team(2023)]{gemini}
Team, G.
\newblock Gemini: a family of highly capable multimodal models.
\newblock \emph{arXiv preprint arXiv:2312.11805}, 2023.

\bibitem[Tian et~al.(2024)Tian, Jiang, Yuan, Peng, and Wang]{tjy+24}
Tian, K., Jiang, Y., Yuan, Z., Peng, B., and Wang, L.
\newblock Visual autoregressive modeling: Scalable image generation via
  next-scale prediction.
\newblock \emph{Advances in neural information processing systems},
  37:\penalty0 84839--84865, 2024.

\bibitem[Tran et~al.(2019)Tran, Tran, Nguyen, Yang, and Cheung]{ttn+19}
Tran, N.-T., Tran, V.-H., Nguyen, B.-N., Yang, L., and Cheung, N.-M.~M.
\newblock Self-supervised gan: Analysis and improvement with multi-class
  minimax game.
\newblock In \emph{Advances in Neural Information Processing Systems}, 2019.

\bibitem[Vaidya(1989)]{v89}
Vaidya, P.~M.
\newblock A new algorithm for minimizing convex functions over convex sets.
\newblock In \emph{30th Annual Symposium on Foundations of Computer Science},
  pp.\  338--343. IEEE Computer Society, 1989.

\bibitem[Valiant \& Valiant(2017)Valiant and Valiant]{valiant2017automatic}
Valiant, G. and Valiant, P.
\newblock An automatic inequality prover and instance optimal identity testing.
\newblock \emph{SIAM Journal on Computing}, 46\penalty0 (1):\penalty0 429--455,
  2017.

\bibitem[Vaswani et~al.(2017)Vaswani, Shazeer, Parmar, Uszkoreit, Jones, Gomez,
  Kaiser, and Polosukhin]{vsp+17}
Vaswani, A., Shazeer, N., Parmar, N., Uszkoreit, J., Jones, L., Gomez, A.~N.,
  Kaiser, {\L}., and Polosukhin, I.
\newblock Attention is all you need.
\newblock \emph{Advances in neural information processing systems}, 30, 2017.

\bibitem[Veličković et~al.(2018)Veličković, Cucurull, Casanova, Romero,
  Liò, and Bengio]{vcc+18}
Veličković, P., Cucurull, G., Casanova, A., Romero, A., Liò, P., and Bengio,
  Y.
\newblock Graph attention networks.
\newblock In \emph{International Conference on Learning Representations}, 2018.
\newblock URL \url{https://openreview.net/forum?id=rJXMpikCZ}.

\bibitem[Wang et~al.(2023)Wang, Lyu, Ji, Zhang, Yu, Shi, and Tu]{wlj+23}
Wang, L., Lyu, C., Ji, T., Zhang, Z., Yu, D., Shi, S., and Tu, Z.
\newblock Document-level machine translation with large language models.
\newblock \emph{EMNLP}, 2023.

\bibitem[Wang et~al.(2022)Wang, Wen, Zhang, Hou, Liu, and Li]{wwz+22}
Wang, X., Wen, K., Zhang, Z., Hou, L., Liu, Z., and Li, J.
\newblock Finding skill neurons in pre-trained transformer-based language
  models.
\newblock \emph{EMNLP}, 2022.

\bibitem[Wei et~al.(2022)Wei, Chen, and Ma]{wcm+22}
Wei, C., Chen, Y., and Ma, T.
\newblock Statistically meaningful approximation: a case study on approximating
  turing machines with transformers.
\newblock \emph{Advances in Neural Information Processing Systems},
  35:\penalty0 12071--12083, 2022.

\bibitem[Wu et~al.(2024{\natexlab{a}})Wu, Hu, Hsiao, and Liu]{whh+24}
Wu, D., Hu, J. Y.-C., Hsiao, T.-Y., and Liu, H.
\newblock Uniform memory retrieval with larger capacity for modern hopfield
  models.
\newblock In \emph{International Conference on Machine Learning}, pp.\
  53471--53514. PMLR, 2024{\natexlab{a}}.

\bibitem[Wu et~al.(2024{\natexlab{b}})Wu, Hu, Li, Chen, and Liu]{whl+24}
Wu, D., Hu, J. Y.-C., Li, W., Chen, B.-Y., and Liu, H.
\newblock {ST}anhop: Sparse tandem hopfield model for memory-enhanced time
  series prediction.
\newblock In \emph{The Twelfth International Conference on Learning
  Representations (ICLR)}, 2024{\natexlab{b}}.

\bibitem[Xie et~al.(2022)Xie, Qiu, Pasad, Du, Qu, and Mei]{xqp+22}
Xie, S., Qiu, J., Pasad, A., Du, L., Qu, Q., and Mei, H.
\newblock Hidden state variability of pretrained language models can guide
  computation reduction for transfer learning.
\newblock \emph{EMNLP}, 2022.

\bibitem[Xu et~al.(2024)Xu, Wang, Li, Wang, Zhao, Chen, Yu, Liu, and
  Wang]{xwl+24}
Xu, H., Wang, S., Li, N., Wang, K., Zhao, Y., Chen, K., Yu, T., Liu, Y., and
  Wang, H.
\newblock Large language models for cyber security: A systematic literature
  review.
\newblock \emph{arXiv preprint arXiv:2405.04760}, 2024.

\bibitem[Xu et~al.(2022)Xu, Zhang, Liu, Sugiyama, and Kankanhalli]{xzl+22}
Xu, X., Zhang, J., Liu, F., Sugiyama, M., and Kankanhalli, M.
\newblock Adversarial attack and defense for non-parametric two-sample tests.
\newblock In \emph{International Conference on Machine Learning}, pp.\
  24743--24769. PMLR, 2022.

\bibitem[Yang et~al.(2019)Yang, Dai, Yang, Carbonell, Salakhutdinov, and
  Le]{ydy+19}
Yang, Z., Dai, Z., Yang, Y., Carbonell, J., Salakhutdinov, R.~R., and Le, Q.~V.
\newblock Xlnet: Generalized autoregressive pretraining for language
  understanding.
\newblock \emph{Advances in neural information processing systems}, 32, 2019.

\bibitem[Yao et~al.(2024)Yao, Li, Pan, and Mei]{ylpm24}
Yao, T., Li, Y., Pan, Y., and Mei, T.
\newblock Hiri-vit: Scaling vision transformer with high resolution inputs.
\newblock \emph{TPAMI}, 2024.

\bibitem[Zamani \& Bendersky(2024)Zamani and Bendersky]{zb24}
Zamani, H. and Bendersky, M.
\newblock Stochastic rag: End-to-end retrieval-augmented generation through
  expected utility maximization.
\newblock \emph{SIGIR}, 2024.

\bibitem[Zandieh et~al.(2021)Zandieh, Han, Avron, Shoham, Kim, and
  Shin]{zha+21}
Zandieh, A., Han, I., Avron, H., Shoham, N., Kim, C., and Shin, J.
\newblock Scaling neural tangent kernels via sketching and random features.
\newblock \emph{Advances in Neural Information Processing Systems},
  34:\penalty0 1062--1073, 2021.

\bibitem[Zelikman et~al.(2023)Zelikman, Huang, Liang, Haber, and
  Goodman]{zhl+23}
Zelikman, E., Huang, Q., Liang, P., Haber, N., and Goodman, N.~D.
\newblock Just one byte (per gradient): A note on low-bandwidth decentralized
  language model finetuning using shared randomness.
\newblock \emph{arXiv preprint arXiv:2306.10015}, 2023.

\bibitem[Zhang(2024)]{zha24}
Zhang, J.
\newblock Graph unlearning with efficient partial retraining.
\newblock In \emph{Companion Proceedings of the ACM on Web Conference 2024},
  pp.\  1218--1221, 2024.

\bibitem[Zhang et~al.(2021)Zhang, Liu, and Zhou]{zlz21}
Zhang, J., Liu, F., and Zhou, A.
\newblock Off-tanet: A lightweight neural micro-expression recognizer with
  optical flow features and integrated attention mechanism.
\newblock In \emph{Pacific Rim International Conference on Artificial
  Intelligence}, pp.\  266--279. Springer, 2021.

\bibitem[Zhang et~al.(2024)Zhang, Xue, Fan, Xu, Li, Pei, and Liu]{zxf+24}
Zhang, J., Xue, R., Fan, W., Xu, X., Li, Q., Pei, J., and Liu, X.
\newblock Linear-time graph neural networks for scalable recommendations.
\newblock In \emph{Proceedings of the ACM on Web Conference 2024}, pp.\
  3533--3544, 2024.

\bibitem[Zhou et~al.(2023)Zhou, Zhu, Chen, Chen, Zhao, Chen, Lin, Wen, and
  Han]{zzc+23}
Zhou, K., Zhu, Y., Chen, Z., Chen, W., Zhao, W.~X., Chen, X., Lin, Y., Wen,
  J.-R., and Han, J.
\newblock Don't make your llm an evaluation benchmark cheater.
\newblock \emph{arXiv preprint arXiv:2311.01964}, 2023.

\end{thebibliography}
\else 

\fi




\end{document}